\documentclass[10pt,journal,compsoc]{IEEEtran}

\ifCLASSOPTIONcompsoc
  \usepackage[nocompress]{cite}
\else
\usepackage{cite}
\fi
\usepackage{booktabs} % For formal tables
\usepackage{xcolor}
\usepackage{soul}
\usepackage{epsfig}
\usepackage{graphicx}
\usepackage{amsmath}
\usepackage{amssymb}
\usepackage{courier}
\usepackage{helvet}
\usepackage{courier}
\usepackage{algorithm}
\usepackage{algorithmic}
\usepackage{multirow}
\usepackage{url}
\usepackage{booktabs}
\usepackage{array}
\usepackage{paralist,algorithmic,algorithm}
\usepackage{balance}
\usepackage[numbers]{natbib}

\newcommand{\x}{{\bf x}}

\newcommand{\y}{{\bf y}}
\newcommand{\X}{{\bf X}}
\newcommand{\D}{\mathcal{D}}

\newcommand{\Y}{\mathcal{Y}}
\newcommand{\R}{\mathbb{R}}
\newcommand{\tr}{\mathop{\mathrm{tr}}}
\newtheorem{defn}{Definition}
\newtheorem{theo}{Theorem}
\newcommand{\tabincell}[2]{\begin{tabular}{@{}#1@{}}#2\end{tabular}}
\newcolumntype{L}[1]{>{\raggedright\let\newline\\\arraybackslash\hspace{0pt}}m{#1}}

\ifCLASSINFOpdf
\else
\fi

\begin{document}
\title{Semi-Supervised Multi-Modal Multi-Instance Multi-Label Deep Network with Optimal Transport}

\author{Yang Yang,~\IEEEmembership{}
        Zhao-Yang Fu,~\IEEEmembership{}
        De-Chuan Zhan,~\IEEEmembership{}
        Zhi-Bin Liu,~\IEEEmembership{}
        and Yuan Jiang,~\IEEEmembership{}
\IEEEcompsocitemizethanks{\IEEEcompsocthanksitem Yang Yang, Zhao-Yang Fu, De-Chuan Zhan and Yuan Jiang are with National Key Laboratory for Novel Software Technology, Nanjing University, Nanjing 210023, China.\protect\\
E-mail: {yangy, fuzy, zhandc, jiangy}@lamda.nju.edu.cn}% <-this % stops an unwanted space
\IEEEcompsocitemizethanks{\IEEEcompsocthanksitem Zhi-Bin Liu is with Tencent WXG, ShenZhen 518057, China.\protect\\
E-mail: lewiszbliu@tencent.com}% <-this % stops an unwanted space
\thanks{De-Chuan Zhan is the corresponding author.}}

\markboth{Journal of \LaTeX\ Class Files,~Vol.~xx, No.~x, xxx~xxx}%
{Shell \MakeLowercase{\textit{et al.}}: Bare Demo of IEEEtran.cls for Computer Society Journals}

\IEEEtitleabstractindextext{%
\begin{abstract}
Complex objects are usually with multiple labels, and can be represented by multiple modal representations, e.g., the complex articles contain text and image information as well as multiple annotations. Previous methods assume that the homogeneous multi-modal data are consistent, while in real applications, the raw data are disordered, e.g., the article constitutes with variable number of inconsistent text and image instances. Therefore, Multi-modal Multi-instance Multi-label (M3) learning provides a framework for handling such task and has exhibited excellent performance. However, M3 learning is facing two main challenges: 1) how to effectively utilize label correlation; 2) how to take advantage of multi-modal learning to process unlabeled instances. To solve these problems, we first propose a novel Multi-modal Multi-instance Multi-label Deep Network (M3DN), which considers M3 learning in an end-to-end multi-modal deep network and utilizes consistency principle among different modal bag-level predictions. Based on the M3DN, we learn the latent ground label metric with the optimal transport. Moreover, we introduce the extrinsic unlabeled multi-modal multi-instance data, and propose the M3DNS, which considers the instance-level auto-encoder for single modality and modified bag-level optimal transport to strengthen the consistency among modalities. Thereby M3DNS can better predict label and exploit label correlation simultaneously. Experiments on benchmark datasets and real world WKG Game-Hub dataset validate the effectiveness of the proposed methods.
\end{abstract}

}

\maketitle

\IEEEdisplaynontitleabstractindextext

\IEEEpeerreviewmaketitle

\IEEEraisesectionheading{\section{Introduction}\label{sec:introduction}}

\begin{figure}[t]
\begin{center}
\begin{minipage}[h]{90mm}
\centering
\includegraphics[width=90mm ]{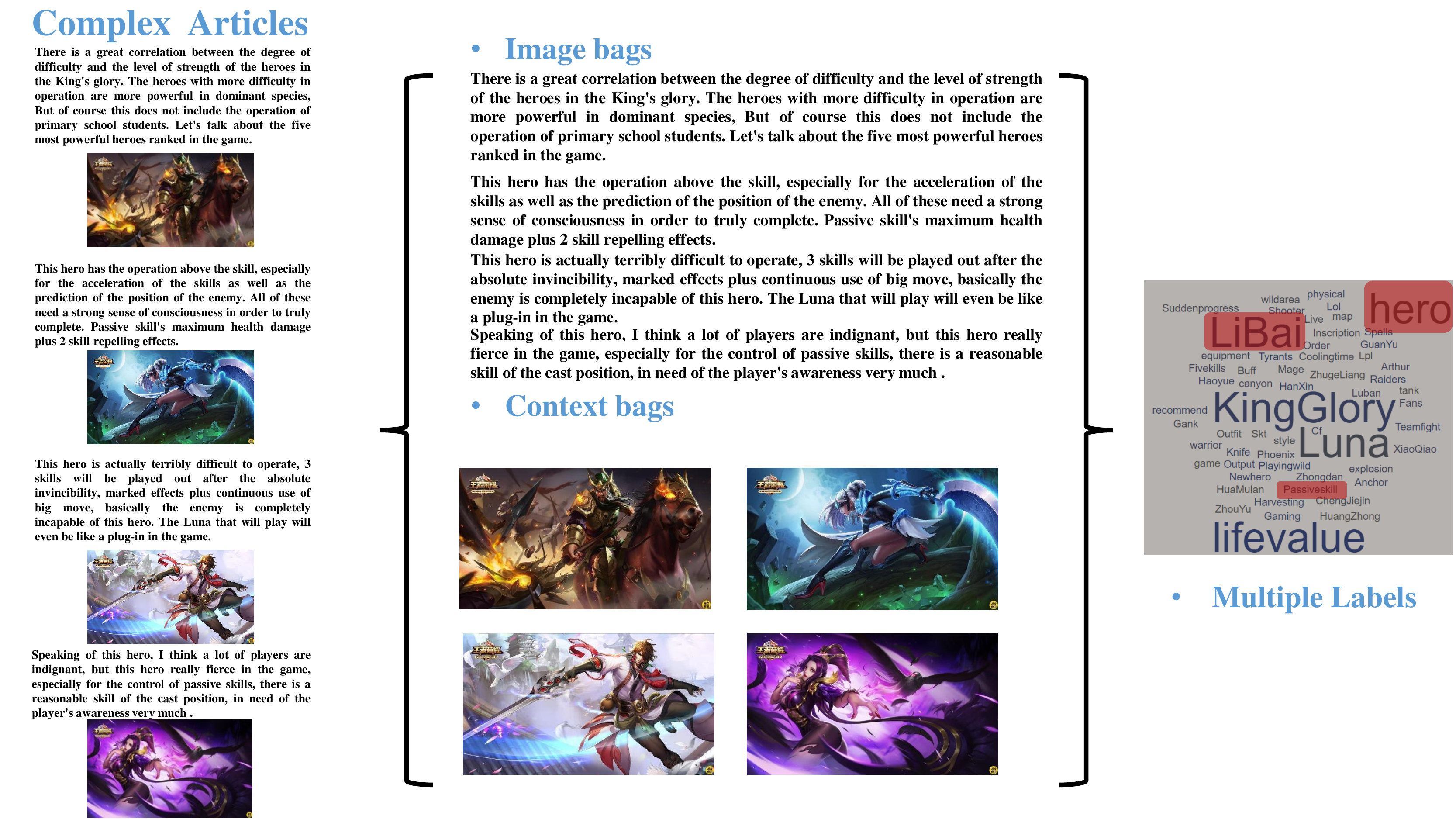}
\end{minipage}
\end{center}
\caption{An illustration of the M3 (Multi-Modal Multi-instance Multi-label) Data in an article of WKG Game-Hub. Each article is with context bag and image bag, each bag contains variable number of instances (context paragraphs/images), while each article has multiple label representations. It is notable that different modalities are heterogeneous, i.e., there have no congruent relationships between the articles and images.}\label{fig:data}
\end{figure}

\IEEEPARstart{W}{ith} the development of data collection techniques, objects can always be represented by multiple modal features, e.g., in the forum of famous mobile game `` Strike of Kings'', the articles are with image and content information, and they belong to multiple categories if they are observed from different aspects, e.g., an article belongs to ``Wukong Sun'' (Game Heroes) as well as ``golden cudgel'' (Game Equipment) from the images, while it can be categorized as ``game strategy'', ``producer name'' from contents and so on. The major challenge for addressing such problem is how to jointly model multiple types of heterogeneities in a mutually beneficial way. To solve this problem, multi-modal multi-label learning approaches utilize multiple modal information, and require modal-based classifiers to generate similar predictions, e.g., \citeauthor{HuangWW15} proposed a multi-label conditional restricted boltzmann machine, which uses multiple modalities to obtain shared representations under the supervision~\cite{HuangWW15}; \citeauthor{YangYFZYLH16} learned a novel graph-based model to learn both label and feature heterogeneities~\cite{YangYFZYLH16}. However, a real-world object may contain variable number of inconsistent multi-modal instances, e.g., the article usually contains multiple images and content paragraphs, in which each image or content paragraph can be regarded as an instance, yet the relationships between the images and contents have not been marked as shown in Figure. \ref{fig:data}.

Therefore, several Multi-modal Multi-instance Multi-label methods have been proposed. \citeauthor{NguyenZZ13} proposed M3LDA with a visual-label part, a textual-label part, and a label topic part, in which the topic decided by visual information and the topic decided by textual information should be consistent~\cite{NguyenZZ13}; \citeauthor{NguyenWLZ14} developed a multi-modal MIML framework based on hierarchical bayesian network~\cite{NguyenWLZ14}. Nevertheless, there are two drawbacks of the existing M3 models. In detail, previous approaches rarely consider the correlations among labels, besides, M3 methods are all supervised methods, which violate the intuition of multi-modal learning using unsupervised data.

Thus, considering the label correlation, \citeauthor{YangH15} studied a hierarchical multi-latent space, which can leverage the task relatedness, modal consistency and the label correlation simultaneously to improve the learning performance~\cite{YangH15}; \citeauthor{HuangZ12} proposed the ML-LOC approach which allows label correlation to be exploited locally~\cite{HuangZ12}; \citeauthor{FrognerZMAP15} developed a loss function with ground metric for multi-label learning, which is based on the wasserstein distance~\cite{FrognerZMAP15}. Previous works mainly assumed that there exists some prior knowledge such as label similarity matrix or the ground metric~\cite{FrognerZMAP15,RoletCP16}. In reality, semantic information among labels is indirect or complicated, thus the confidence of the label similarity matrix or ground metric is weak. On the other hand, considering the labeling cost, there are many unlabeled instances. The most important advantage of multi-modal methods is that they use unlabeled data, e.g., co-training~\cite{BlumM98} style methods utilized the complementary principle to label unlabeled data for each other; co-regularize~\cite{BrefeldGSW06} style methods exploited unlabeled multi-modal data with consistency principle. Meanwhile, it is notable that previous proposed M3 based methods are hard to adopt the unlabeled instances. Therefore, another issue is how to bypass the limitation of M3 style methods by using unlabeled multi-modal instances.

%\begin{figure}[t]
%\centering
%\includegraphics[width=90mm,height=70mm]{data.pdf}
%\caption{An illustration of the M3 (Multi-Modal Multi-instance Multi-label) Data in real-world application as article of WKG Game-Hub. Each article is with context bag and image bag as multiple modalities, each bag contains variable number of instances (context paragraphs/images), while each article has multiple label representations. It is notable that different modalities are heterogeneous, i.e., there have no congruent relationships between the articles and images.}\label{fig:data}
%\end{figure}

In this work, aiming at learning the label prediction and exploring label correlation with semi-supervised M3 data simultaneously, we proposed a novel general Multi-modal Multi-instance Multi-label Deep Network, which models the independent deep network for each modality, and imposes the modal consistency on bag-level prediction. To better consider the label correlation, M3DN first adopts Optimal Transport (OT)~\cite{villani2008optimal}  distance to measure the quality of prediction. The adoption provides a more meaningful measure in multi-label tasks by capturing the geometric information of the underlying label space. The raw data may not calculate the raw ground metric confidently, thus we cast the label correlation exploration as a latent ground metric learning problem. Moreover, considering the unlabeled data information, we propose the semi-supervised M3DN (M3DNS). M3DNS utilizes the instance-level auto-encoder to build the single modal network, and considers the bag-level consistency among different unlabeled modal predictions with the modified OT theory. Consequently, M3DNS could automatically learn the predictors from different modalities and the latent shared ground metric.

The main contributions of this paper are summarized in the following points:
\begin{itemize}
\item We propose a novel Multi-modal Multi-instance Multi-label Deep Network (M3DN), which models the deep independent network for each modality, and imposes the modal consistency on bag-level prediction;
\item We consider label correlation exploration as a latent ground metric learning problem between different modalities, rather than a fix ground metric using prior raw knowledge;
\item We utilize the extrinsic unlabeled data, by considering instance-level auto-encoder, and the bag-level consistency among different unlabeled modal predictions with the modified OT metric;
\item We achieve superior performances on real-world applications, comprehensively evaluate on the performance and obtain consistently superior performances stably.
\end{itemize}

Section 2 summarizes related work, our approaches are presented in Section 3. Section 4 reports our experiments. Finally, Section 5 gives the conclusion.

\section{Related Work}
\IEEEPARstart{T}{he} exploitation of multi-modal multi-instance multi-label learning has attracted much attention recently. In this paper, our method concentrates on deep multi-label classification for semi-supervised inconsistent multi-modal multi-instance data, and considers the label correlation using optimal transport technique. Therefore, our work is related to M3 learning and the optimal transport.

Multi-modal learning deals with data from multiple modalities, i.e., multiple feature sets. The goals are to improve performance and reduce the sample complexity. Meanwhile, multi-modal multi-label learning has been well studied, e.g., \citeauthor{FangZ12} proposed a multi-modal multi-label learning method based on the large margin framework~\cite{FangZ12}. \citeauthor{YangHYF14} modeled both the modal consistency and the label correlation in a graph-based framework~\cite{YangHYF14}. The basic assumption behind these methods is that multi-modal data is consistent. However, in real applications, the multi-modal data are always heterogeneous on the instance-level, e.g., articles have variable number of inconsistent images and text paragraphs, videos have variable length of inconsistent audio and image frames. Articles and videos only have consistency on the bag level, rather than instance level. Thus, multi-modal multi-instance multi-label learning is proposed recently. \citeauthor{NguyenWLZ14} developed a multi-modal MIML framework based on hierarchical bayesian network~\cite{NguyenWLZ14}; \citeauthor{FengZ17} exploited deep neural network to generate instance representation for MIML and it can be extended to multi-modal scenario. Nevertheless, previous approaches rarely consider the confidence of label correlation. More importantly, the current M3 approaches are supervised, which obviously lose the advantage of multi-modal learning for processing unlabeled data.

Considering the label correlation, several multi-label learning methods are proposed~\cite{BiK14,ZhangZ14,ZhanZ17}. Recently, Optimal Transport (OT) ~\cite{villani2008optimal} is developed to measure the difference between two distributions based on given ground metric, and it has been widely used in computer vision and image processing fields, e.g., \citeauthor{qian2016non} proposed a novel method that exploits knowledge in both data manifold and feature correlation~\cite{qian2016non}; \citeauthor{CourtyFTR17} proposed a regularized unsupervised optimal transportation model to perform the alignment of the representations~\cite{CourtyFTR17}. However, previous works mainly assumed that prior knowledge for cost matrix already exists,  and ignored deficiency of information or domain knowledge. Thus, \citeauthor{CuturiA14, zhao2018} suggested to formulate the cost metric learning problem with the side information~\cite{CuturiA14, zhao2018}. On the other hand, existing M3 methods are almost supervised methods, while multi-modal methods aim to utilize the complementary~\cite{BlumM98} or consistency~\cite{BrefeldGSW06} principle using the unlabeled instance. Thereby how to take unlabeled data into consideration becomes a challenge.
%
%Therefore, the main contribution of this paper is a novel deep M3 deep framework, which considers label correlation better and the semi-supervised extension.
%Therefore, to solve these problems, we proposed a novel Multi-modal Multi-instance Multi-label Deep Network (M3DN) and the semi-supervised M3DN (M3DNS), which learn the label prediction and exploring label correlation simultaneously, and can improve the label prediction performance as a result. Specifically, M3DN inputs bag of instances to different modal deep networks, based on Optimal Transport (OT) theory, bag-level label predictions corresponding to ground truth labels and extra modal predictions, are required consistent with the learned label metric. On the other hand, considering the uncertainty of the prior knowledge, we cast the label correlation exploration as a latent ground metric learning problem. Based on M3DN, M3DNS considers both the instance-level and bag-level unlabeled modal instances, which can learn more discriminative models and further improve the prediction performance. Consequently, M3DN and M3DNS could automatically learn the predictors for different modalities and the latent shared ground metric.

\section{Proposed Method}
\subsection{Notation}
\IEEEPARstart{I}{n} the multi-instance extension of the multi-modal multi-label framework, we are given $N$ bags of instances, let $\Y = \{\y_1,\y_2,\cdots,\y_{N_l}\}$ denotes the label set, $\y_i \in \R^L$ is the label vector of $i-$th bag, where $y_{i,j} = 1$ denotes positive class, and $y_{i,j} = 0$ otherwise. On the other hand, suppose we are given $K$ modalities, without any loss of generality, we consider two modalities in our paper, i.e., images and contents. Let $\D = \{([\X_1^1,\X_1^2],\y_1),([\X_2^1,\X_2^2],\y_2), \cdots , ([\X_{N_l}^1,\X_{N_l}^2],\y_{N_l}),\\([\X_{N_l+1}^1,\X_{N_l+1}^2]),\cdots,([\X_{N_l+N_u}^1,\X_{N_l+N_u}^2])\}$ represents the training dataset, where $N_l/N_u$ denotes the number of labelled/unlabelled instances. $\X_i^1 = \{\x_{i,1}^1, \x_{i,2}^1, \cdots,\x_{i,m_i}^1\}$ denotes the bag representation of $m_i$ instances of $\X_i^1$, similarly, $\X_i^2 = \{\x_{i,1}^2,\x_{i,2}^2, \cdots,\x_{i,n_i}^2\}$ is the bag representation of $n_i$ instances of $\X_i^2$, it is notable that bags of different modalities may contain variable number of instances.

The goal is to generate a learner to annotate new bags based on its inputs $\X^1, \X^2$, e.g., annotate a new complex article with its images and contents.

\begin{figure}[t]
\begin{center}
\begin{minipage}[h]{90mm}
\centering
\includegraphics[width=90mm ]{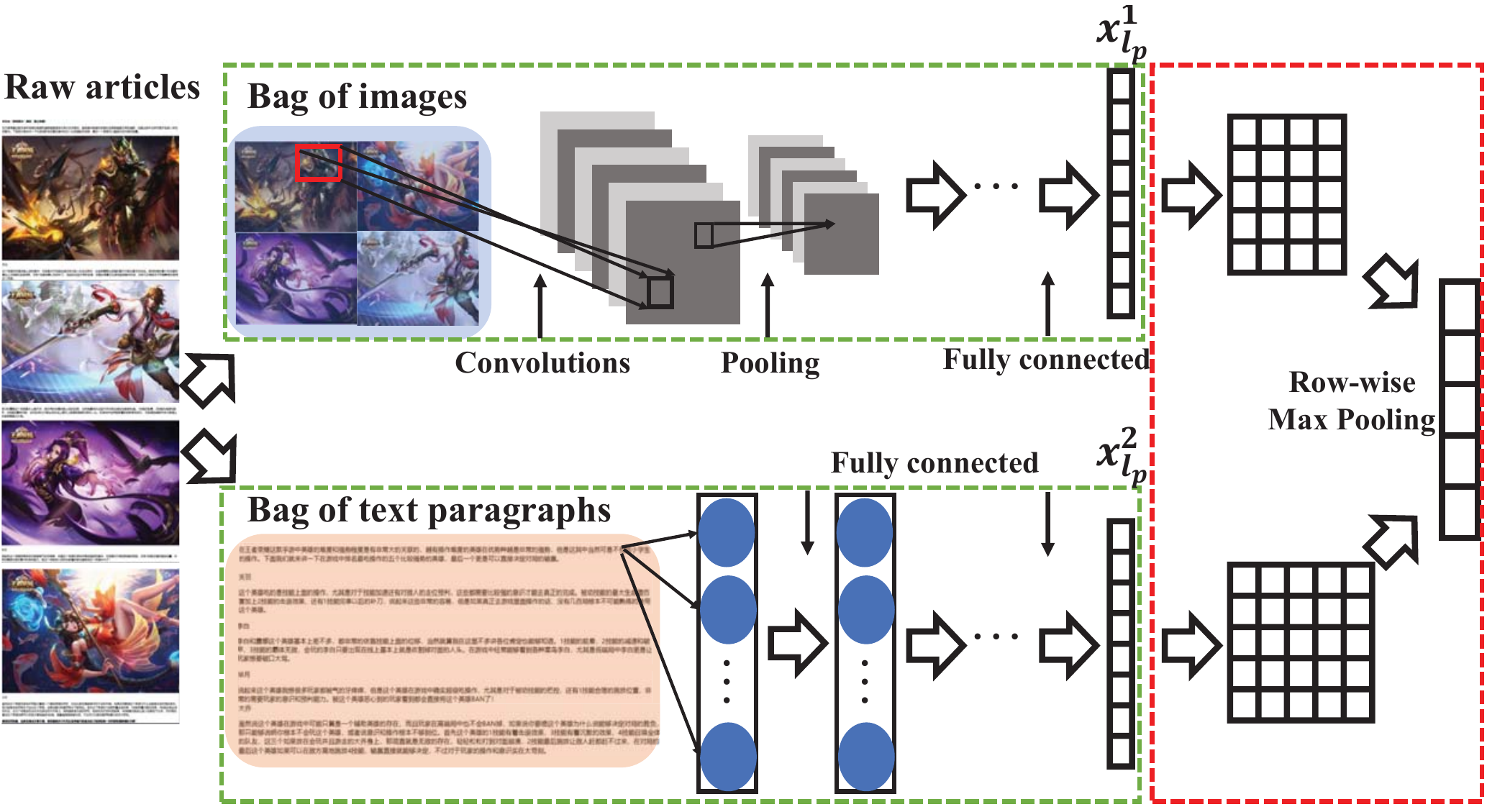}
\end{minipage}
\end{center}
\caption{The flowchart of the M3DN, the raw articles can be divided into two homogeneous modal bag with variable number of heterogeneous instances, i.e., the image bag with four images and content bag with 5 text paragraphs. The instances of different modalities can be calculated with different deep networks, and finally represented as $\x_{l_p}^1$ or $\x_{l_p}^2$, the output features are fully connected with the labels, and we can get the bag-concept layer for different modalities. Eventually, we can acquire the final prediction by mean-max pooling the bag-concept layer of different modalities.}\label{fig:f1}
\end{figure}

%\begin{figure*}[t]
%\centering
%\includegraphics[width=150mm,height=70mm]{framework.pdf}
%\caption{The flowchart of the M3DN, the raw articles can be divided into two homogeneous modal bag with variable number of heterogeneous instances, i.e., the image bag with four images and content bag with 5 text paragraphs. The instances of different modalities can be calculated with different deep networks, and finally represented as $\x_{l_p}^1$ or $\x_{l_p}^2$, the output features are fully connected with the labels, and we can get the bag-concept layer for different modalities. Eventually, we can acquire the final prediction by mean-max pooling the bag-concept layer of different modalities.}\label{fig:f1}
%\end{figure*}

\subsection{Optimal Transport}
Traditionally, several measurements such as Kullback-Leibler divergences, Hellinger and total variation, have been utilized to measure the similarity between two distributions. However, these measurements play little effect when the probability space has geometrical structures. On the other hand, Optimal transport~\cite{villani2008optimal}, also known as Wasserstein distance or earth mover distance~\cite{yossi1997earth}, defines a reasonable distance between two probability distribution over the metric space. Intuitively, the Wasserstein distance is the minimum cost of transporting the pile of one distribution into the pile of another distribution, which formulates the problem of learning the ground metric as minimizing the difference between two polyhedral convex functions over a convex set of distance matrices. Therefore, the Wasserstein distance is more powerful in such situations by considering the pairwise cost.

\noindent\begin{defn}\label{de:d2}
(Transport Polytope) For two probability vectors $r$ and $c$ in the simplex $\sum_L$, $U(r, c)$ is the transport polytope of $r$ and $c$, namely the polyhedral set of $L \times L$ matrices,
\begin{equation}
U(r, c) = \{P\in \R_+^{L \times L} | P{\bf 1}_L = r, P^\top{\bf 1}_L = c\} \nonumber
\end{equation}
\end{defn}

\noindent\begin{defn}\label{de:d1}
(Optimal Transport) Given a $L \times L$ cost matrix $M$, the total cost of mapping from $r$ to $c$ using a transport matrix (or coupling probability) $P$  can be quantified as $\langle P,M \rangle$. The optimal transport (OT) problem is defined as,
\begin{equation}
d_M(r,c) = \min\limits_{P \in U(r,c)}\langle P,M \rangle \nonumber
\end{equation}
\end{defn}

When $M$ belongs to the cone of metric matrices $\mathbb{M}$, the value of $d_M(r,c)$ is a distance~\cite{villani2008optimal} between $r$ and $c$, parameterized by $M$. In that case, assuming implicitly that $M$ is fixed and only $r$ and $c$ vary, we will refer to the optimal transport distance between $r$ and $c$. It is notable that $d_M(r,c)$ is the cost of the optimal plan for transporting the predicted mass distribution $r$ to match the target distribution $c$. The penalty increases when more mass is transported over longer distances, according to the ground metric $M$.

\noindent\begin{theo}
$d_M$ defined in Def. \ref{de:d1} is a distance on $\sum_L$ whenever $M$ is a metric matrix~\cite{villani2008optimal}.
\end{theo}

\subsection{Multi-Modal Multi-instance Multi-label Deep Network (M3DN)}
Multi-modal Multi-instance Multi-label (M3) learning provides a framework for handling the complex objects, and we propose a novel M3 based parallel deep network (M3DN). Based on the M3DN, we can bypass the limitation of initial label correlation metric using the Optimal Transport (OT) theory, and further take advantage of unlabeled data considering the modal consistency. In this section, we propose the Multi-Modal Multi-instance Multi-label Deep Network (M3DN) framework. M3DN models deep networks for different modalities and imposes the modal consistency.

\begin{figure}[t]
\begin{center}
\begin{minipage}[h]{85mm}
\centering
\includegraphics[width=85mm ]{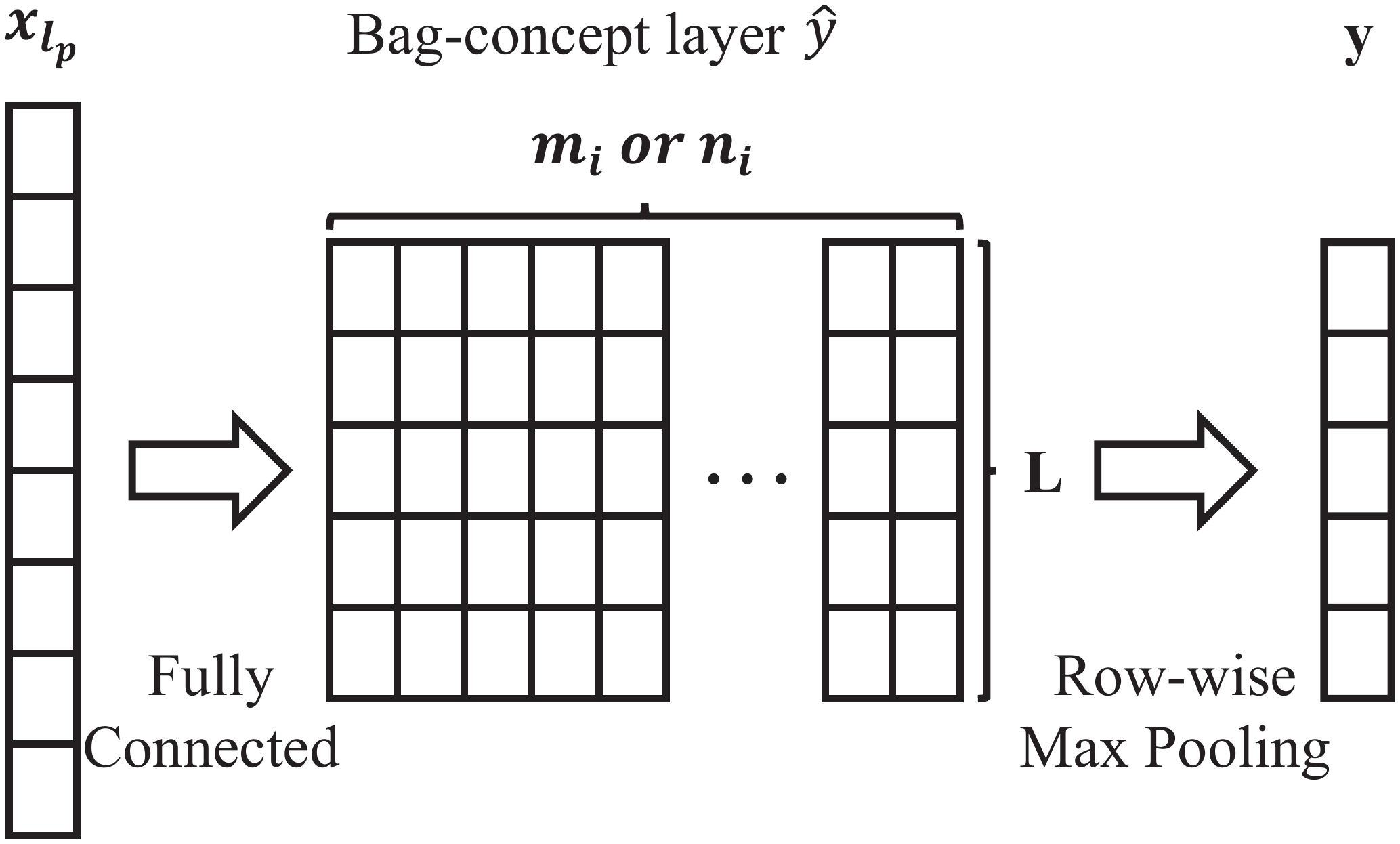}
\end{minipage}
\end{center}
\caption{The schematic of the bag-concept layer. We can acquire the bag-concept layer with the output feature representations of a bag of instances, in which each column represents corresponding prediction of each instance. Eventually, the final label prediction is calculated by row-wise max pooling.}\label{fig:f2}
\end{figure}
The raw articles contain variable number of heterogeneous multi-modal information, i.e., when no corresponding relationships exist among each the contents and images, it is difficult to utilize the consistency principle with previous multi-modal methods. Thus, we turn to utilize the consistency among the bags of different modalities, rather than the instance-level. Specifically, raw articles can be divided into two modal bags of heterogeneous instances, i.e., the image bag with 4 images and content bag with 5 text paragraphs as shown in Fig. \ref{fig:f1}, while only the homogeneous bags share the same multiple labels. Each instance $\x^1(\x^2)$ in different modal bag can be calculated among several layers and can be finally represented as $\x_{l_{p1}}(\x_{l_{p2}})$.

Without any loss of generality, we use the convolutional neural network for images and the fully connected networks for text. Then, the output features are fully connected with the bag-concept layer. All parameters including deep network facts and fully connected weights can be organized as $\Theta_1=\{\theta_{l_1},\theta_{l_2},\cdots,\theta_{l_{p1-1}},W_1\}(\Theta_2=\{\theta_{l_1},\theta_{l_2},\cdots,\theta_{l_{p2-1}},W_2\})$. Concretely, once the label predictions of the instances for a bag $\X_i^v$ are obtained, we propose a fully connected 2D layer (bag-concept layer) with the size of $m_i(n_i) \times L$ as shown in Fig. \ref{fig:f2}, in which each column represents corresponding prediction of each instance in the image/content bag. Formally, for a given bag of instances $\X_i^v$, the $(k, j)$-th node in the 2D bag-concept layer represents the prediction score between the instance $\x_{i,j}^v$ and the $k-$th label. Therefore, the $j$-column has the following form of activation:
\begin{equation}\label{eq:output}
\hat{\y}_{j}^v = g(W_{v}\x_{i,j}^v + b_{v})
\end{equation}
Here, $g(\cdot)$ can be any convex activation function, and we use softmax function here. In the bag-concept layer, we utilize the row-wise max pooling: $f_v(i) = max(\hat{\y}_{i,\cdot})$. The final prediction value is:$f = \frac{f_1+f_2}{2}$.

\subsection{Explore Label Correlation}

However, fully connection to the label output rarely considers the relationship among labels. Recently, Optimal Transport (OT) theory ~\cite{villani2008optimal} is used in multi-label learning, which captures geometric information of the underlying label space. According to the Def. ~\ref{de:d1} and Def. ~\ref{de:d2}, the loss function implied in the parallel network structure can be formulated without any loss of generality as:
\begin{equation}\label{eq:e1}\small
\begin{split}
&\min\limits_{P_v \in U(f(X_i^v),\y_i)} {\sum_{v = 1}^2\sum_{i=1}^N}\langle P_v,M \rangle\\
s.t.  \quad & U(f(\X_i^v),\y_i) = \{P_v\in \R_+^{L \times L} | P_v{\bf 1}_L = f(X_i^v), P_v^\top{\bf 1}_L = \y_i\}
\end{split}
\end{equation}
where $M$ is the shared latent cost matrix. However, this method requires prior knowledge to construct the cost matrix $M$. However, in reality, indirect or incomplete information among labels leads to weak cost matrix $M$ and poor classification performance.

Therefore, we can define the process of learning cost metric as an optimization problem. Optimizing the cost metric directly is difficult and it consumes $O(L^2)$ constraints. Thus, ~\cite{CuturiA14, zhao2018} proposed to formulate the cost metric learning problem with the side information, i.e., the label similarity matrix $S$ as~\cite{zhao2018}, and~\cite{CuturiA14} has proved that the cost metric matrix $M$, which computes corresponding optimal transport distance $d_M$ between pairs of labels, agrees with the side information. More precisely, this criterion favors matrix $M$, in which the distance $d_M(r;c)$ is small for pairs of similar histograms $r$ and $c$ (corresponding $S(r;c)$ is large) and large for pairs of dissimilar histograms (corresponding $S(r;c)$ is small). Consequently, optimizing $M$ can be turned to optimize the $S$. Finally, the goal of M3DN can be turned to learn label predictor and explore label correlation simultaneously.

In detail, we first introduce the connection between nonlinear transformation and pseudo-metric:
\noindent\begin{defn}\label{de:d3}
With the nonlinear transformation $\emptyset(\cdot)$, the Euclidean distance after the transformation can be denoted as:
\begin{equation}
\begin{split}
D_\emptyset(r,c) = \|\emptyset(r) - \emptyset(c)\|_2. \nonumber
\end{split}
\end{equation}
And \cite{KedemTWSL12} proved that $D_\emptyset$ satisfies all properties of a well-defined pseudo-metric in the original input space.
\end{defn}

\noindent\begin{theo}
For a pseudo-metric $M$ defined in Def. 3 and histograms $r, c \in \sum_L$, the function $(r, c) \rightarrow {\bf 1}_{r\neq c}d_M(r, c)$ satisfies all four distance axioms, i.e., non-negativity, symmetry, definiteness and sub-additivity (triangle inequality) as in~\cite{CuturiA14}.
\end{theo}

Thus, $M$ can be turned to learn the kernel $S$ defined by the non-linear transformation $\emptyset(\cdot)$:
\begin{equation}\label{eq:ker}
\begin{split}
S_{ij} = S(\y_i, \y_j ) = \emptyset(\y_i)^\top \emptyset(\y_j)
\end{split}
\end{equation}
where the $\y_i$ represents the label vector of $i-$th instance. Besides, it is notable that the cost matrix $M$ is computed as $M_{ij} = D_\emptyset^2(\y_i, \y_j)$, while the kernel $S$ is defined as Eq.~\ref{eq:ker}. Thus, the relation between $M$ and $S$ can be derived as:
\begin{equation}\label{eq:e3}
M_{ij} = S_{ii} + S_{jj} - 2S_{ij}.
\end{equation}
The non-linear mapping preserves pseudo metric properties in Def. 3, therefore it only needs a projection to positive semi-definite matrix cone when learning the kernel matrix $S$. Thus, we can avoid the projection to metric space which is complicated and costly. Therefore, we propose to conduct the label predictions and label correlation exploration simultaneously based on substituted optimal transport, the combination of Eq. ~\ref{eq:e3} and Eq.~\ref{eq:e1} can be reformulated as:
\begin{equation}\label{eq:e2}\small
\begin{split}
&\min\limits_{S, P_v \in U(f(X_i^v),\y_i)} {\sum_{v = 1}^2\sum_{i=1}^N
}\langle P_v,M \rangle + \lambda_1 r(S,S_0)\\
s.t.  \quad &U(f(\X_i^v),\y_i) = \{P_v\in \R_+^{L \times L} | P_v{\bf 1}_L = f(X_i^v), P_v^\top{\bf 1}_L = \y_i\}\\
 \quad & S \in \mathcal{S}_{+}, \quad M_{ij} = S_{ii} + S_{jj} - 2S_{ij}
\end{split}
\end{equation}
where $\lambda_1$ is a trade-off parameter, $\mathcal{S}_{+}$ denotes the set of positive semi-definite matrix. We adopt OT distance as the loss between prediction and groundtruth, and then incorporate the ground metric learning by kernel biased regularization in 2nd term, where $\lambda_1 r(S,S_0)$ can be any convex regularization. The regularizer $\mathcal{S}_{+} \times \mathcal{S}_{+} \rightarrow \mathcal{R}_{+}$ allows us to exploit prior knowledge on the kernelized similar matrix, encoded by a reference matrix $S_0$. Since typically no strong prior knowledge is available, we use $S_0 = \Y' \times \Y$. Following common practice~\cite{HoffmanRDKS14}, we utilize the asymmetric Burg divergence, which yields:
\begin{equation}
r(S,S_0) = \tr(SS_0^{-1}) - log det(SS_0^{-1}) - p \nonumber
\end{equation}
where $p$ is the balance parameter, and we set as 1 in our experiments.

\begin{figure}[t]
\begin{center}
\begin{minipage}[h]{90mm}
\centering
\includegraphics[width=90mm ]{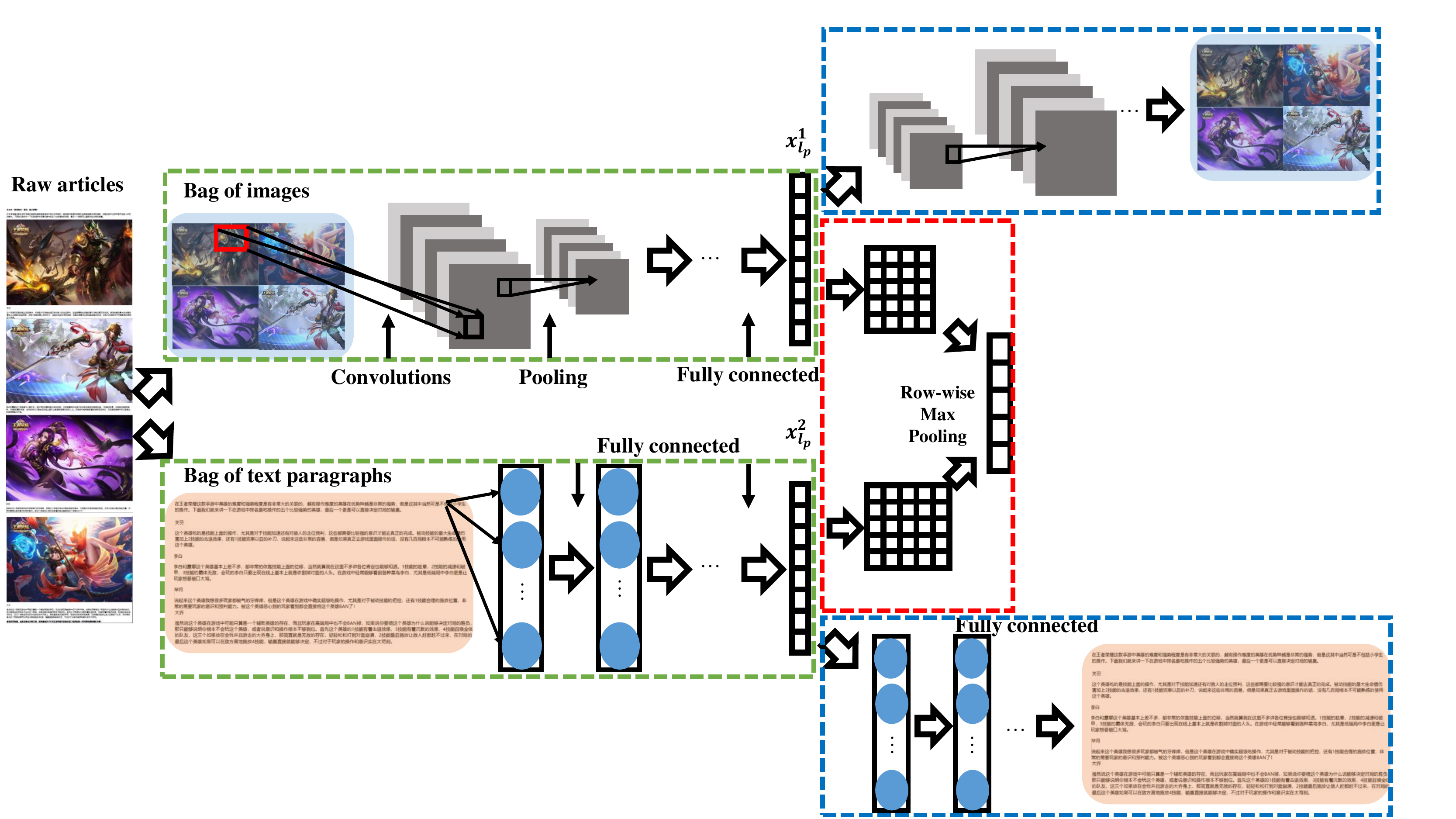}
\end{minipage}
\end{center}
\caption{The flowchart of the M3DNS consider unlabeled data. Similar to M3DN, the raw articles can be divided into two homogeneous modal bags with variable number of heterogeneous instances. The instances of different modalities can be calculated with different deep networks, and finally represented as $\x_{l_p}^1$ or $\x_{l_p}^2$. The output features of labeled data are fully connected with the labels, while we add decoder networks for each modality to process the unlabeled data. On the other hand, we can get bag representations of all data from the bag-concept layer for different modalities. Eventually, we can acquire the final predictions of different modalities and calculate the semi-supervised loss.}\label{fig:msdns}
\end{figure}

\subsection{Consider Unsupervised Data}
M3DN provides a framework for handling complex multi-modal multi-instance multi-label objects, and it considers the label correlation as an optimization problem in Eq. \ref{eq:unloss}. The limitation of manual labeling is that, in real application, it leaves over large number of unlabeled data. In other words, unlabeled data is readily available, while labeled data tends to be of smaller size. The basic intuition of multi-modal learning is to utilize the complement or consistent information of unlabeled data, to get better performance. Yet M3DN leaves the unlabeled data without consideration, and this obviously loses the advantage of multi-modal learning. Consequently, how to extend M3DN to semi-supervised scenario is an urgent problem.

To consider the extrinsic consistency, i.e., the unlabeled information of different modalities, we propose a semi-supervised M3DN (M3DNS) methods for learning each modal predictors. Different from previous co-regularize style methods using instance-level consistency principle, M3 learning only has bag-level consistency among different modalities, rather than instance-level consistency. Thus, there exist two challenges in using unlabeled data in M3 learning: 1) how to utilize different modal instance-level unlabeled data; 2) how to utilize different modal bag-level consistency of unlabeled data.

To solve this problem, M3DNS utilizes the instance-level unlabeled instances with auto-encoder and bag-level unlabeled instances with modified OT. As shown in Fig. \ref{fig:msdns}, since different modal bags include various number of instances, and the correspondences among different modal instances are unknown, we turn to utilize the auto-encoder based networks to reconstruct the input instances for different modalities, which can build more robust encoder networks. On the one hand, bag-level correspondences are known, thereby for the bag-level unlabeled data, we utilize modified OT consistency term to constraint different modalities.

Specifically, each modal ordinal network can be replaced by auto-encoder (AE) network, which minimizes the reconstruction error of all the instances, i.e., auto-encoder CNN for image modality and auto-encoder fully connected network for content modality. Without any loss of generality, AE can be formulated as square loss:
\begin{equation}\label{eq:ae}
\begin{split}
AE(\x_k) = \min\limits_{\Theta_{f_v},\Theta_{r_v}} \sum_{i=N_l+1}^{N_u}\|\x_{i^v} - r_v(f_v(x_{i^v}))\|_F^2 \\
\end{split}
\end{equation}
where $\Theta_{f_v}, \Theta_{r_v}$ are the weight parameters of encoder network $f_v$ and decoder network $r_v$ of the $v-$th modality.

On the other hand, Eq. \ref{eq:e1} only utilizes the supervised information, while neglect the unlabeled modal bag-level correspondences. Thus, with the unlabeled information, Eq. \ref{eq:e1} can be reformulated as:
\begin{equation}\label{eq:unlabel}\small
\begin{split}
&\min\limits_{P_v \in U, \hat{P} \in \hat{U}} {\sum_{v = 1}^2\sum_{i=1}^{N_l}}\langle P_v,M \rangle + \sum_{i=1}^{N_u} \langle \hat{P},M \rangle\\
s.t.  \quad & U = \{P_v\in \R_+^{L \times L} | P_v{\bf 1}_L = f(X_i^v), P_v^\top{\bf 1}_L = \y_i\}\\
\quad & \hat{U} = \{\hat{P}\in \R_+^{L \times L} | \hat{P}{\bf 1}_L = f(X_i^1), \hat{P}^\top{\bf 1}_L = f(X_i^2)\}
\end{split}
\end{equation}
where $\hat{P}$ is the pseudo transport matrix (or coupling probability) for unlabeled data. The extra unlabeled modal predictions can be regarded as the pseudo labels in $\hat{P}$ for constructing more discriminative predictors. In detail, when learning one modal predictor, the predictions of other modalities can act as the pseudo label, which can assist learning more discriminative predictors with unlabeled data. Thus M3DNS can well utilize the bag-level consistency among different modalities. Therefore, M3DNS can acquire more robust ground metric $M$, which potentially utilizes the consistency between different modal bags.

As a result, with the unlabeled information, we can combine the Eq. \ref{eq:unlabel} and Eq. \ref{eq:ae}. The semi-supervised M3DN method (M3DNS) can be given as:
\begin{equation}\label{eq:unloss}
\begin{split}
\min\limits_{P_v \in U, \hat{P} \in \hat{U}} &\sum_{v = 1}^2\sum_{i=1}^{N_l}\langle P_v,M \rangle + \sum_{i=N_l+1}^{N_u}AE(x_i^v) + \sum_{i=1}^{N_u} \langle \hat{P},M \rangle\\
& + \lambda_1 r(S,S_0)\\
s.t.  \quad & U = \{P_v\in \R_+^{L \times L} | P_v{\bf 1}_L = f(X_i^v), P_v^\top{\bf 1}_L = \y_i\}\\
\quad & \hat{U} = \{\hat{P}\in \R_+^{L \times L} | \hat{P}{\bf 1}_L = f(X_i^1), \hat{P}^\top{\bf 1}_L = f(X_i^2)\}\\
\quad & S \in \mathcal{S}_{+}, \quad M_{ij} = S_{ii} + S_{jj} - 2S_{ij}
\end{split}
\end{equation}

\subsection{Optimization}

The $\hat{P}$ is similar with the $P$ when considering the extra modal predictions as the pseudo label. Thus, we analyze the optimization of the Eq. \ref{eq:e2}, and Eq. \ref{eq:unloss} has similar solution. In detail, The 1st term in Eq. \ref{eq:e2} involves the product of predictors $f$ and cost matrix $S$, which makes the formulation not joint convex. Consequently, the formulation cannot be optimized easily. We provide the optimization process below:

{\begin{algorithm}[t]
\caption{The pseudo code of learning the predictors}
\label{alg:alg1}
\leftline{\bf Input:}
\begin{compactitem}
 \item Sampled Batch Dataset: $\{[X_i^1, X_i^2],\y\}_{i=1}^n$, kernelized similar matric $S^t$, current mapping $f_1, f_2$
 \item Parameter: $\lambda$
\end{compactitem}
\leftline{\bf Output:}
\begin{compactitem}
\item Gradient of the target mapping: $\partial{L}/\partial{f_1}, \partial{L}/\partial{f_2}$
\end{compactitem}
\begin{algorithmic}[1]{
\STATE Calculate $M \leftarrow $ Eq.~\ref{eq:e3}
\STATE Initialize $K = exp(-\lambda M-1)$, $\nabla \leftarrow$ {\bf 0}
\FOR {$v = 1,2$}
\FOR {$i = 1,2,\cdots,n$}
\STATE $u_i^v \leftarrow$ {\bf 1}
\WHILE {$u_i^v$ not converged}
\STATE $u_i^v \leftarrow f_v(\x_i^v)\oslash(K(\y_i^v\oslash K^\top u_i^v))$
\ENDWHILE
\STATE $\nabla^{f_v} \leftarrow \nabla^{f_v} + \frac{log u_i^v}{\lambda} - \frac{log {u_i^v}^\top {\bf 1}}{\lambda L}\cdot{\bf 1}$
\ENDFOR
\ENDFOR}
\end{algorithmic}
\end{algorithm}}

{\bf Fix $S$, Optimize $f_1,f_2$:}
When updating $f_1, f_2$ with a fixed $S$, the 2nd term of Eq.~\ref{eq:e2} is irrelevant to $f_1, f_2$, and the Eq.~\ref{eq:e2} can be reformulated as follows:
\begin{equation}\label{eq:e4}\small
\begin{split}
& \min\limits_{P_v \in U(f(X_i^v),\y_i)} {\sum_{v = 1}^2\sum_{i=1}^N}\langle P_v,M \rangle \\
s.t. \quad & U(f(\X_i^v),\y_i) = \{P_v\in \R_+^{L \times L} | P_v{\bf 1}_L = f(X_i^v), P_v^\top{\bf 1}_L = \y_i\}\\
\end{split}
\end{equation}
The empirical risk minimization function of Eq.~\ref{eq:e4} can be optimized by stochastic gradient descent. However, it requires to evaluate the descent direction for the loss, with respect to the predictor $f$. Computing the exact subgradient is quite costly, it needs to solve a linear program with $O(L^2)$ constraints, which are with high expense with the $L$ (the label dimension) increase.

Similar to~\cite{FrognerZMAP15}, the loss is a linear program, and the subgradient can be computed using Lagrange duality. Therefore, we use primal-dual approach to compute the gradient by solving the dual LP problem. From~\cite{bertsimas1997introduction}, we know that the dual optimal $\alpha$ is, in fact, the subgradient of the loss of training sample $(\X^v, \y)$ with respect to its first argument $f_v$. However, it is costly to compute the exact loss directly. In~\cite{Cuturi13}, Sinkhorn relaxation is adopted as the entropy regularization to smooth the transport objective, which results in a strictly convex problem that can be solved through Sinkhorn matrix scaling algorithm, at a speed that is faster than that of transport solvers~\cite{Cuturi13}.

For a given training bag of instances $([\X^1, \X^2], \y)$, the dual LP of Eq.~\ref{eq:e4} is:
\begin{equation}\label{eq:e5}
\begin{split}
d_M(f_v(\X^v),y) = \max\limits_{\alpha,\beta \in C_M} \alpha^\top f(\X_i^v) + \beta \y,
\end{split}
\end{equation}
where $C_M = \{\alpha,\beta \in \R^L : \alpha_i + \beta_j < M_{i,j}\}$.

\noindent\begin{defn}\label{de:d4}
(Sinkhorn Distance) Given a $L \times L$ cost matrix $M$, and histograms $(r,c) \in \sum_L$. The Sinkhorn distance is defined as:
\begin{equation}\label{eq:e6}
\begin{split}
&d_M^\lambda(r,c) = \min\limits_{P^\lambda \in U(r,c)}\langle P^\lambda,M \rangle \\
&P^\lambda = \arg\min\limits_{P \in U(f(X_i^v),\y_i)} \langle P,M \rangle - \frac{1}{\lambda}H(P)
\end{split}
\end{equation}
\end{defn}
where $H(P) = -\sum_{i=1}^L\sum_{j=1}^L p_{ij} log p_{ij}$ is the entropy of $P$, and $\lambda > 0$ is entropic regularization coefficient.

\begin{table*}[t]{\scriptsize
\centering
\caption{Comparison results (mean $\pm$ std.) of M3DN/M3DNS with compared methods on benchmark datasets.}
\label{tab:tab11}
\begin{tabular*}{1\textwidth}{@{\extracolsep{\fill}}@{}l|c|@{}c|@{}@{}c@{}@{\;}|c|c|c|c|c@{}}
\toprule
\multirow{2}{*}{Methods} & \multicolumn{4}{c|}{Coverage $\downarrow$} & \multicolumn{4}{c}{Macro AUC $\uparrow$}\\
\cmidrule(l){2-9}
& FLICKR25K & IAPR TC-12 & MS-CoCo & NUS-WIDE & FLICKR25K & IAPRTC-12 & MS-CoCo & NUS-WIDE \\
\midrule
M3LDA  & 12.345$\pm$.214 & 11.620$\pm$.042 & 47.400$\pm$.622 & 6.670$\pm$.205 & .532$\pm$.015	& .526$\pm$.003	& .507$\pm$.015 & .509$\pm$.012 \\
MIMLmix & 17.114$\pm$1.024 & 15.720$\pm$.543 &64.130$\pm$1.121 &14.167$\pm$1.140 & .472$\pm$.018 & .554$\pm$.096 &.471$\pm$.019 &.493$\pm$.020 \\
\midrule
CS3G  &8.168$\pm$.137 &7.153$\pm$.178 &50.138$\pm$2.146 &8.028$\pm$.907 & .837$\pm$.007 & .817$\pm$.006 &.717$\pm$.011 &.530$\pm$.022\\ \midrule
DeepMIML & 9.242$\pm$.331 & 8.931$\pm$.421  & 27.358$\pm$.654 & 8.369$\pm$.119 & .766$\pm$.035 &  .795$\pm$.022 & .827$\pm$0.006 & .823$\pm$.005 \\
M3MIML &11.760$\pm$1.121 &9.125$\pm$.553 &42.420$\pm$.2.696 &5.210$\pm$.920 &.687$\pm$.087 &.724$\pm$.033 &.650$\pm$.032 &.649$\pm$.084\\
MIMLfast&12.155$\pm$.913 &12.711$\pm$.315 &41.048$\pm$.831 &8.634$\pm$.028 &.524$\pm$.050 &.485$\pm$.009 &.506$\pm$.010 &.522$\pm$.008\\ \midrule
SLEEC &9.568$\pm$.222 &9.494$\pm$.105 &47.502$\pm$.448 &7.390$\pm$.275 &.706$\pm$.007 &.675$\pm$.007 &.661$\pm$.014 &.620$\pm$.006\\
Tram &7.959$\pm$.187 &8.156$\pm$.163 &28.417$\pm$.945 &9.934$\pm$.026 &.780$\pm$.009 &.746$\pm$.007 &.776$\pm$.011 &.493$\pm$.007\\
ECC &14.818$\pm$.086 &14.229$\pm$.258 & 47.124$\pm$.675 &7.941$\pm$.194 &.532$\pm$.013 &.484$\pm$.009 &.630$\pm$.023 &.634$\pm$.009\\
ML-KNN &10.379$\pm$.115 &9.523$\pm$.072 &27.568$\pm$.066 &4.610$\pm$.062 &.591$\pm$.008 &.723$\pm$.006 & .823$\pm$.003 &.736$\pm$.008\\
RankSVM &11.439$\pm$.196 &11.941$\pm$.078 &37.300$\pm$.835 &8.292$\pm$.054 &.512$\pm$.019 &.499$\pm$.009 &.521$\pm$.033 &.501$\pm$.001\\
ML-SVM &11.311$\pm$.158 &11.755$\pm$.270 &39.258$\pm$.294 &7.890$\pm$.020 &.503$\pm$.010 &.502$\pm$.010 &.497$\pm$.016 &.561$\pm$.001\\ \midrule
M3DN & 7.502$\pm$.129 & 6.936$\pm$.065 & 26.921$\pm$.320 & 4.599$\pm$.050 &.822 $\pm$.009 &.798$\pm$.002  & .811$\pm$.004  & .826$\pm$.006 \\
M3DNS &\bf 3.947$\pm$.307 &\bf 4.214$\pm$.202 &\bf 6.119$\pm$.262 &\bf 2.764$\pm$.071 &\bf.892$\pm$.004 &\bf .876$\pm$.003  &\bf .838$\pm$.003  &\bf .898$\pm$.008 \\
\end{tabular*}
\begin{tabular*}{1\textwidth}{@{\extracolsep{\fill}}@{}l|c|c|c|c|c|c|c|c@{}}
\toprule
\multirow{2}{*}{Methods} & \multicolumn{4}{c|}{Ranking Loss $\downarrow$} & \multicolumn{4}{c}{Example AUC $\uparrow$}\\
\cmidrule(l){2-9}
& FLICKR25K & IAPR TC-12 & MS-CoCo & NUS-WIDE & FLICKR25K & IAPRTC-12 & MS-CoCo & NUS-WIDE \\
\midrule
M3LDA & .301$\pm$.009 & .377$\pm$.002 & .247$\pm$.001 & .257$\pm$.006 & .707$\pm$.008 & .630$\pm$.005 & .770$\pm$.006 & .652$\pm$.009 \\
MIMLmix & .609$\pm$.036 & .675$\pm$.012 &.609$\pm$.040 &.583$\pm$.081 & .391$\pm$.036 & .325$\pm$.012 &.391$\pm$.040 &.417$\pm$.082 \\
\midrule
CS3G  &.118$\pm$.005 &.155$\pm$.005 &.202$\pm$.009 &.170$\pm$.032 &.881$\pm$.005 &.835$\pm$.005 &.798$\pm$.009 &.642$\pm$.032\\ \midrule
DeepMIML & .149$\pm$.012 & .166$\pm$.017 & .089$\pm$.002 & .164$\pm$.007 & .791$\pm$.044 & .834$\pm$.017 & .911$\pm$.002 & .835$\pm$.007 \\
M3MIML &.271$\pm$.053 &.250$\pm$.011 &.191$\pm$.016 &.284$\pm$.030 &.729$\pm$.053 &.751$\pm$.011 &.811$\pm$.017 &.717$\pm$.031\\
MIMLfast&.275$\pm$.033 &.435$\pm$.021 &.194$\pm$.006 &.430$\pm$.009 &.724$\pm$.033 &.626$\pm$.013 &.811$\pm$.005 &.646$\pm$.009\\ \midrule
SLEEC &.316$\pm$.009 &.413.006 &.455$\pm$.005 &.512$\pm$.008 &.843$\pm$.003 &.761$\pm$.005 &.796$\pm$.002 &.713$\pm$.008\\
Tram &.132$\pm$.004 &.203$\pm$.007 &.117$\pm$.004 &.456$\pm$.004 &.867$\pm$.004 &.797$\pm$.007 &.883$\pm$.005 &.591$\pm$.001\\
ECC &.804$\pm$.024 &.928$\pm$.013 &.461$\pm$.009 &.617$\pm$.020 &.642$\pm$.005 &.529$\pm$.012 &.775$\pm$.005 &.697$\pm$.013\\
ML-KNN &.235$\pm$.005 &.264$\pm$.004 &.097$\pm$.002 &.176$\pm$.003 &.764$\pm$.005 &.736$\pm$.004 &.903$\pm$.001 &.824$\pm$.003\\
RankSVM &.236$\pm$.006 &.344$\pm$.001 &.199$\pm$.098 &.323$\pm$.008 &.763$\pm$.006 &.656$\pm$.001 &.801$\pm$.098 &.677$\pm$.001\\
ML-SVM &.232$\pm$.005 &.337$\pm$.009 &.179$\pm$.004 &.314$\pm$.002 &.768$\pm$.005 &.662$\pm$.009 &.822$\pm$.004 &.686$\pm$.002\\ \midrule
M3DN & .108$\pm$.003 & .151$\pm$.002 &\bf .085$\pm$.002 &\bf .117$\pm$.002 & .891$\pm$.003 & .850$\pm$.003 &\bf .915$\pm$.003 &\bf .883$\pm$.001 \\
M3DNS &\bf .108$\pm$.001 &\bf .142$\pm$.002 & .112$\pm$.003 & .119$\pm$.003 &\bf .899$\pm$.004 &\bf .858$\pm$.005  & .898$\pm$.008  & .881$\pm$.006 \\
\end{tabular*}
\begin{tabular*}{1\textwidth}{@{\extracolsep{\fill}}@{}l|c|c|c|c|c|c|c|c@{}}
\toprule
\multirow{2}{*}{Methods} & \multicolumn{4}{c|}{Average Precision $\uparrow$} & \multicolumn{4}{c}{Micro AUC $\uparrow$}\\
\cmidrule(l){2-9}
& FLICKR25K & IAPR TC-12 & MS-CoCo & NUS-WIDE & FLICKR25K & IAPRTC-12 & MS-CoCo & NUS-WIDE \\
\midrule
M3LDA  & .371$\pm$.005 & .311$\pm$.007 & .399$\pm$.007 & .338$\pm$.005 & .693$\pm$.006 & .609$\pm$.002 & .773$\pm$.005 & .657$\pm$.008\\
MIMLmix & .207$\pm$.038 & .183$\pm$.008 &.213$\pm$.041 &.167$\pm$.020 & .436$\pm$.024 & .438$\pm$.060 &.434$\pm$.026 &.472$\pm$.015 \\
\midrule
CS3G  &\bf .749$\pm$.008 &.622$\pm$.006 &.542$\pm$.012 &.597$\pm$.031 &.867$\pm$.005 &.827$\pm$.006 &.738$\pm$.007 &.557$\pm$.021\\ \midrule
DeepMIML & .621$\pm$.027 & .619$\pm$.025 & .633$\pm$.005 & .583$\pm$.008 & .835$\pm$.009 & .802$\pm$.017 & .914$\pm$.002 & .852$\pm$.003 \\
M3MIML &.423$\pm$.056 &.490$\pm$.020 &.446$\pm$.030 &.443$\pm$.076 &.745$\pm$.034 &.707$\pm$.017 &.816$\pm$.020 &.762$\pm$.020\\
MIMLfast&.432$\pm$.064 &.339$\pm$.013 &.413$\pm$.005 &.365$\pm$.021 &.712$\pm$.022 &.540$\pm$.010 &.745$\pm$.012 &.630$\pm$.005\\ \midrule
SLEEC &.608$\pm$.006 &.473$\pm$.010 &.565$\pm$.003 &.392$\pm$.007 &.824$\pm$.004 &.736$\pm$.005 &.795$\pm$.002 &.701$\pm$.005\\
Tram &.653$\pm$.011 &.523$\pm$.008 &.494$\pm$.007 &.336$\pm$.002 &.842$\pm$.003 &.782$\pm$.007 &.883$\pm$.006 &.554$\pm$.002\\
ECC &.416$\pm$.012 &.278$\pm$.011 &.462$\pm$.007 &.438$\pm$.014 &.646$\pm$.004 &.514$\pm$.008 &.779$\pm$.005 &.702$\pm$.009\\
ML-KNN &.398$\pm$.006 &.403$\pm$.010 &.585$\pm$.002 &.439$\pm$.006 &.752$\pm$.005 &.729$\pm$.003 &.905$\pm$.002 &.817$\pm$.004\\
RankSVM &.467$\pm$.005 &.364$\pm$.004 &.427$\pm$.066 &.401$\pm$.001 &.748$\pm$.005 &.649$\pm$.004 &.791$\pm$.093 &.680$\pm$.003\\
ML-SVM &.466$\pm$.006 &.367$\pm$.006 &.441$\pm$.007 &.443$\pm$.007 &.753$\pm$.004 &.656$\pm$.009 &.825$\pm$.004 &.724$\pm$.001\\ \midrule
M3DN &.719$\pm$.006 & .634$\pm$.003 & .680$\pm$.005 &\bf .691$\pm$.001 &\bf .876$\pm$.003 & .834$\pm$.001  &\bf .918$\pm$.002 & .877$\pm$.003 \\
M3DNS & .698$\pm$.002 &\bf .637$\pm$.007 &\bf .691$\pm$.004 & .634$\pm$.003 &.858$\pm$.003 &\bf .863$\pm$.004  & .877$\pm$.006  &\bf .878$\pm$.005 \\
\bottomrule
\end{tabular*}}
\end{table*}

\begin{table*}[t]\centering{\footnotesize
\caption{Comparison results (mean $\pm$ std.) of M3DN/M3DNS with compared methods on WKG Game-Hub dataset.  6 commonly used criteria are evaluated. The best performance for each criterion is bolded. $\uparrow/\downarrow$ indicates the larger/smaller the better of the criterion.}
  \label{tab:tab2}
  \begin{tabular*}{1\textwidth}{@{\extracolsep{\fill}}@{}l|c|c|c|c|c|c}
    \toprule
    \multirow{2}{*}{Methods} & \multicolumn{6}{c}{Content Modality} \\
    \cmidrule(l){2-7}
    & \tabincell{c}{Coverage $\downarrow$ \\ ($\times 10^2$)} & \tabincell{c}{Macro \\ AUC $\uparrow$}  & \tabincell{c}{Ranking \\ Loss $\downarrow$} & \tabincell{c}{Example \\ AUC $\uparrow$} & \tabincell{c}{Average \\ Precision $\uparrow$} &\tabincell{c}{Micro \\ AUC $\uparrow$}\\
    \midrule
    M3LDA & .466$\pm$.020 & .470$\pm$.015 & 1.000$\pm$1.000 & .360$\pm$.056 & .098$\pm$.001 & .381$\pm$.036\\
    MIMLmix & .334$\pm$.003 & .507$\pm$.002 & .445$\pm$.006 & .539$\pm$.001 & .111$\pm$.001 & .540$\pm$.003\\
    \midrule
    CS3G & .362$\pm$.002 & .593$\pm$.001 & .340$\pm$.003 & .659$\pm$.003 & .371$\pm$.002 & .614$\pm$.007\\
    \midrule
    DeepMIML & .341$\pm$.010 & .533$\pm$.018 & .415$\pm$.027 & .186$\pm$.025 & .600$\pm$.030 & .634$\pm$.014\\
	M3MIML &N/A &N/A &N/A &N/A &N/A &N/A \\
	MIMLfast & .363$\pm$.040 &.496$\pm$.050	&.414$\pm$.056	&.585$\pm$.056	&.162$\pm$.033	&.567$\pm$.040\\
    \midrule
    M3DN & .258$\pm$.006 & .761$\pm$.016 & .276$\pm$.008 & .723$\pm$.008 & .329$\pm$.002 & .753$\pm$.007\\
    M3DNS &\bf .246$\pm$.002 &\bf .763$\pm$.001 &\bf .255$\pm$.002 &\bf .744$\pm$.002 &\bf .332$\pm$.001 &\bf .763$\pm$.001\\
    \bottomrule
    \end{tabular*}
    \begin{tabular*}{1\textwidth}{@{\extracolsep{\fill}}@{}l|c|c|c|c|c|c}
    \multirow{2}{*}{Methods} & \multicolumn{6}{c}{Image Modality} \\
    \cmidrule(l){2-7}
    & \tabincell{c}{Coverage $\downarrow$ \\ ($\times 10^2$)} & \tabincell{c}{Macro \\ AUC $\uparrow$}  & \tabincell{c}{Ranking \\ Loss $\downarrow$} & \tabincell{c}{Example \\ AUC $\uparrow$} & \tabincell{c}{Average \\ Precision $\uparrow$} &\tabincell{c}{Micro \\ AUC $\uparrow$}\\
    \midrule
    M3LDA & .466$\pm$.010 & .455$\pm$.054 & 1.000$\pm$.000 & .359$\pm$.019 & .098$\pm$.001 & .384$\pm$.030\\
    MIMLmix & .329$\pm$.002 & .502$\pm$.003 & .427$\pm$.005 & .557$\pm$.001 & .114$\pm$.001 & .560$\pm$.002\\
    \midrule
    CS3G & .395$\pm$.004 & .545$\pm$.001 & .405$\pm$.003 & .595$\pm$.003 & .304$\pm$.003 & .563$\pm$.006\\
    \midrule
    DeepMIML & .383$\pm$.006 & .512$\pm$.002 & .515$\pm$.009 & .484$\pm$.009 & .121$\pm$.001 & .488$\pm$.018\\
	M3MIML &N/A &N/A &N/A &N/A &N/A &N/A \\
	MIMLfast & .402$\pm$.070 & .512$\pm$.061 & .433$\pm$.059 & .566$\pm$.059 & .170$\pm$.037 & .547$\pm$.058\\
    \midrule
    M3DN & .175$\pm$.001 & .896$\pm$.001 & .210$\pm$.002 & .789$\pm$.002 & .402$\pm$.001 & .586$\pm$.000\\
    M3DNS &\bf .164$\pm$.001 &\bf .910$\pm$.003 &\bf .196$\pm$.001 &\bf .803$\pm$.001 &\bf .407$\pm$.000 &\bf .869$\pm$.000\\
    \bottomrule
    \end{tabular*}
  \begin{tabular*}{1\textwidth}{@{\extracolsep{\fill}}@{}l|c|c|c|c|c|c}
    \multirow{2}{*}{Methods} & \multicolumn{6}{c}{Overall} \\
    \cmidrule(l){2-7}
    & \tabincell{c}{Coverage $\downarrow$ \\ ($\times 10^2$)} & \tabincell{c}{Macro \\ AUC $\uparrow$}  & \tabincell{c}{Ranking \\ Loss $\downarrow$} & \tabincell{c}{Example \\ AUC $\uparrow$} & \tabincell{c}{Average \\ Precision $\uparrow$} &\tabincell{c}{Micro \\ AUC $\uparrow$}\\
    \midrule
    M3LDA & .466$\pm$.008 & .468$\pm$.026 & 1.000$\pm$.000 & .359$\pm$.030  & .098$\pm$.001 & .383$\pm$.017 \\
    MIMLmix &.358$\pm$.003 &.504$\pm$.002 &.488$\pm$.007 &.496$\pm$.001 &.101$\pm$.001 &.519$\pm$.003\\
    \midrule
    CS3G &.361$\pm$.004 &.589$\pm$.003 &.346$\pm$.004 &.653$\pm$.004 &.365$\pm$.001 &.612$\pm$.004\\
    \midrule
    DeepMIML & .362$\pm$.005 & .518$\pm$.002 & .488$\pm$.008 & .512$\pm$.008 & .125$\pm$.001 & .524$\pm$.018\\
	M3MIML &N/A &N/A &N/A &N/A &N/A &N/A \\
	MIMLfast& .393$\pm$.060 & .509$\pm$.064 & .430$\pm$.052 & .596$\pm$.052 & .170$\pm$.036 & .549$\pm$.054 \\
    \midrule
    SLEEC & .603$\pm$.013 & .518$\pm$.004 & .756$\pm$.007 & .493$\pm$.005 & .150$\pm$.006 & .583$\pm$.006\\
    Tram & .712$\pm$.005 & .429$\pm$.008 & .109$\pm$.010 & .545$\pm$.003 & .164$\pm$.008 & .464$\pm$.006\\
    ECC & .622$\pm$.017 &.630$\pm$.002 &.632$\pm$.009 &.530$\pm$.017 &.198$\pm$.002 &.592$\pm$.011\\
    ML-KNN &.675$\pm$.020 &.712$\pm$.006 &.175$\pm$.003 &.802$\pm$.015 &.265$\pm$.004 &.814$\pm$.001\\
    RankSVM &N/A &N/A &N/A &N/A &N/A &N/A\\
    ML-SVM & .742$\pm$.023 & .561$\pm$.002 & .223$\pm$.009 & .782$\pm$.008 & .234$\pm$.003 & .793$\pm$.002\\
    \midrule
    M3DN & .163$\pm$.003 & .924$\pm$.002 & .190$\pm$.004 & .809$\pm$.004 & .401$\pm$.003 & .866$\pm$.003\\
    M3DNS &\bf .149$\pm$.002 &\bf .933$\pm$.001 &\bf .180$\pm$.009 &\bf .828$\pm$.003 &\bf .409$\pm$.001 &\bf .880$\pm$.001\\
    \bottomrule
    \end{tabular*}}
\end{table*}

Based on the Sinkhorn theorem, we conclude that the transportation matrix can be written in the form of $P^\star = diag(u) K diag(v)$, where $K = exp(-\lambda M-1)$ is the element-wise exponential of $\lambda M-1$. Besides, $u = exp(\lambda \alpha)$ and $v = exp(\lambda \beta)$.

Therefore, we adopt the well-known Sinkhorn-Knopp algorithm, which is used in~\cite{Cuturi13,CuturiA14} to update the target mapping $f_v$ given the ground metric. $f_v$ can be defined as Eq.~\ref{eq:output}.
% \begin{equation}\label{eq:e7}
% \begin{split}
% &\hat{\y}_i = f(\x_i) = [\hat{y}_1,\hat{y}_2,\cdots,\hat{y}_L]\\
% &\hat{y}_k = \frac{exp(W_{v}\x_{i,j}^v + b_{v})}{\sum_{k=1}^L exp(W_{v}\x_{i,j} + b_{v})}
% \end{split}
% \end{equation}
The detailed procedure is summarized in Algorithm \ref{alg:alg1}, then with the help of Back Propagation technique, gradient descent could be adopted to update the network parameters.

{\begin{algorithm}[t]
\caption{The pseudo code of M3DN} \label{alg:alg2}
\leftline{\bf Input:}
\begin{compactitem}
 \item Dataset: $\D = \{[X_i^1, X_i^2],\y\}_{i=1}^N$
 \item Parameter: $\lambda_1$, $\lambda$
 \item $\rm maxIter$: $T$, learning rate: $\{\alpha_t\}_{t=1}^T$
\end{compactitem}
\leftline{\bf Output:}
\begin{compactitem}
\item Classifiers: $f_1, f_2$
\item Label similar matric: $S, M$
\end{compactitem}
\begin{algorithmic}[1]{
\STATE Initialize $S_0 \leftarrow \Y' \times \Y$
\WHILE {true}
\STATE Create Batch: Randomly pick up $n$ examples from $\D$ without replacement
\STATE Calculate $S^{t+1} \leftarrow $ Eq. \ref{eq:e8}, Eq. \ref{eq:e9}
\STATE Calculate $\partial{L}/\partial{f_1^t}, \partial{L}/\partial{f_2^t} \leftarrow $ Alg. \ref{alg:alg1}
\STATE Weight Propagation step: Obtain the derivative $\partial{f_1^t}/\partial{\Theta_1}$, $\partial{f_2^t}/\partial{\Theta_2}$;
\STATE Update parameters ${\Theta_1}, {\Theta_2}$
\STATE $Func_{obj}^{t+1} \leftarrow$ calculate obj. value in Eq. \ref{eq:e2} with $F^{t+1}$
\IF {$\|Func_{obj}^{t+1} - Func_{obj}^t\|\leq \epsilon$ or $t \geq T$}
\STATE Break;
\ENDIF
\ENDWHILE}
\end{algorithmic}
\end{algorithm}}

{\bf Fix $f_1,f_2$, Optimize $S$:}

When updating $S$ with the fixed $f_1,f_2$, the sub-problem can be rewritten as following:
\begin{equation}\label{eq:e8}
\begin{split}
&\min\limits_{S} {\sum_{v = 1}^2\sum_{i=1}^N}\langle P,M \rangle + \lambda_1 r(S,S_0)\\
s.t. &\quad K \in \mathcal{S}_+,\quad M_{ij} = S_{ii}+S_{jj}-2S_{ij}.\\
\end{split}
\end{equation}
This sub-problem has closed-form solution. The differential can be formulated as:
\begin{equation}\label{eq:e8}
\begin{split}
S = (\bar{P} + S_0^{-1} - p)^{-1}
\end{split}
\end{equation}
where
\begin{equation}
\begin{split}
  \bar{P} = \left\{
\begin{aligned}
&-2P_{ij} ,{\rm ~~~~~~~~~~~~}  {when\quad i \neq j},\\
&\sum_{k \neq i}^L (P_{ik} + P_{ki}) , {when\quad i = j}   \nonumber
\end{aligned}
\right.
\end{split}
\end{equation}

Then, we project $S$ back to positive semi-definite cone as:
\begin{equation}\label{eq:e9}
\begin{split}
S = {\bf Proj}(S) = U max(\sigma, 0)U^\top
\end{split}
\end{equation}
where {\bf Proj} is a projection operator, U and $\sigma$ correspond to the eigenvectors and eigenvalues of $S$. The whole procedure is summarized in Algorithm \ref{alg:alg2}.

Eq. \ref{eq:unloss} can be easily optimized as M3DN with GCD method. Without any loss of generality, in semi-supervised scenario, the extra modal prediction $f(X^{3-i})$ can be regarded as the pseudo label similar to the $\y$ in the supervised term when updating $f_1, f_2$. $S$ can be updated in similar form, where
\begin{equation}\small
\begin{split}
  \bar{P} = \left\{
\begin{aligned}
&-2(P_{ij}+\hat{P}_{ij}) ,{\rm ~~~~~~~~~~~~~~~~~~~~}  {when\quad i \neq j},\\
&\sum_{k \neq i}^L (P_{ik} + P_{ki}+ \hat{P}_{ik} + \hat{P}_{ki}) , {when\quad i = j}   \nonumber
\end{aligned}
\right.
\end{split}
\end{equation}

\begin{figure}[t]
\begin{center}
\begin{minipage}[h]{75mm}
\centering
\includegraphics[width=75mm ]{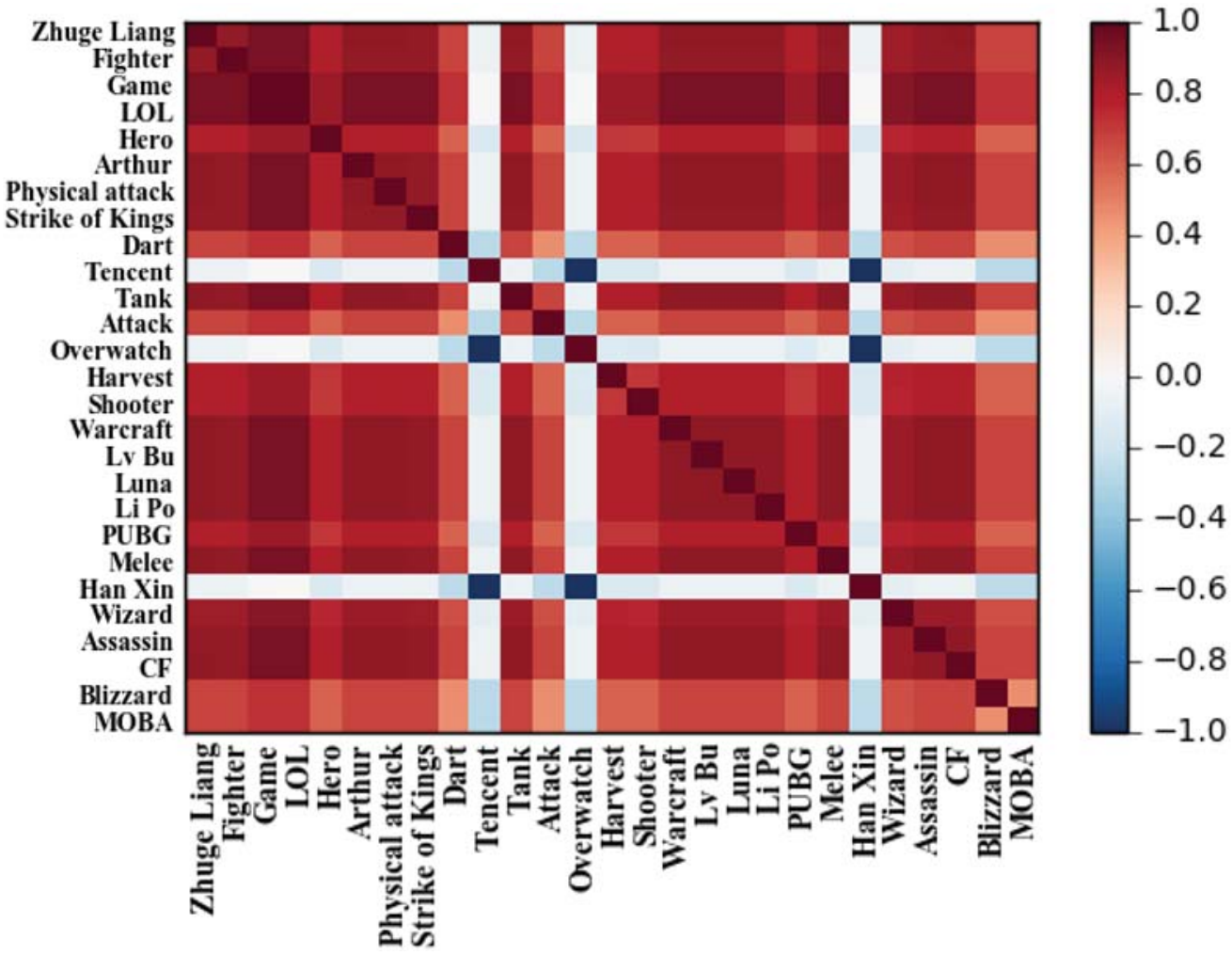}
\end{minipage}
\end{center}
\caption{Illustration of learned label correlations for different datasets, and the value has been scaled in [-1,1]. Red color
indicates a positive correlation, and blue one indicates a negative correlation.}\label{fig:res2}
\end{figure}

\begin{figure*}[t]
\begin{center}
\begin{minipage}[h]{65mm}
\centering
\includegraphics[width=65mm ]{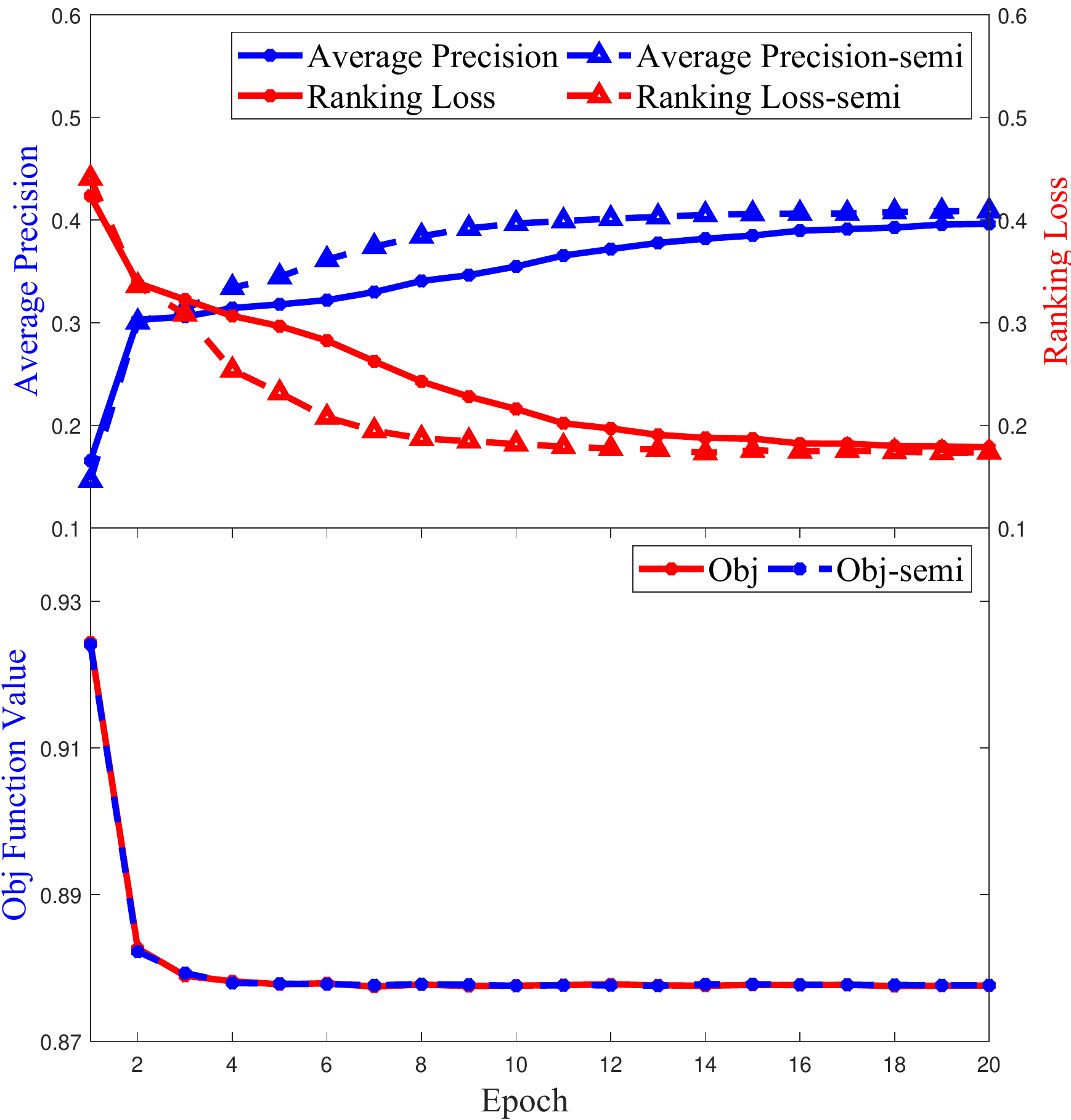}
\mbox{ \;\;\;\; ({\it a}) {M3DN}}
\end{minipage}
\begin{minipage}[h]{65mm}
\centering
\includegraphics[width=65mm ]{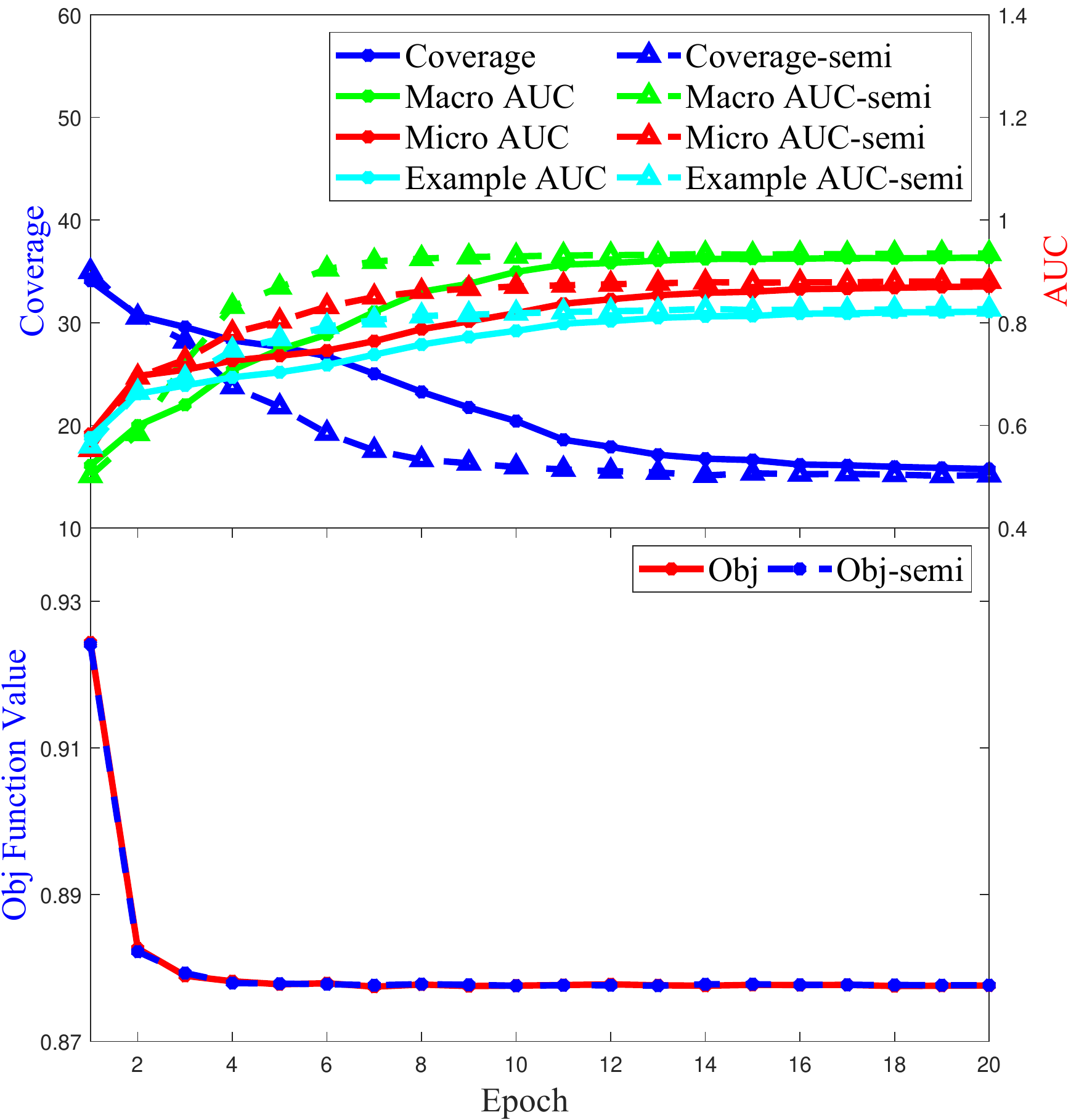}
\mbox{ \;\;\;\; ({\it b}) {M3DNS}}
\end{minipage}
\end{center}
\caption{Objective function value convergence and corresponding classification performance (Coverage, Ranking Loss, Average Precision, Macro AUC, example AUC and Micro AUC) vs. number of iterations of M3DN and M3DNS}\label{fig:convergence}
\end{figure*}

\section{Experiments}

\subsection{Datasets and Configurations}
M3DN/M3DNS can learn more discriminative multi-modal feature representation on bag level for supervised/semi-supervised multi-label classification, while considering the label correlation among different labels. Thus, in this section, we provide empirical investigations and performance comparisons of M3DN on multi-label classification and label correlation. Without any loss of generality,  we experiment on 4 public real-world datasets, i.e., FLICKR25K~\cite{HuiskesL08}, IAPR TC-12~\cite{EscalanteHGLMMSPG10}, MS-COCO~\cite{LinMBHPRDZ14} and NUS-WIDE~\cite{ChuaTHLLZ09}. Besides, we experiment on 1 real-world complex article dataset, i.e., WKG Game-Hub. {\bf FLICKR25K}: consists of 25,000 images collected from Flickr website, and each image is associated with several textual tags. The text for each instance is represented as a 1386-dimensional bag-of-words vector. Each point is manually annotated with 24 labels. We select 23,600 image-text pairs that belong to the 10 most frequent concepts; {\bf IAPR TC-12}: consists of 20,000 image-text pairs which annotate 255 labels. The text for each point is represented as a 2912-dimensional bag-of-words vector; {\bf NUS-WIDE}: contains 260,648 web images, and images are associated with textual tags where each point is annotated with 81 concept labels. We select 195,834 image-text pairs that belong to the 21 most frequent concepts. The text for each point is represented as a 1000-dimensional bag-of-words vector; {\bf MS-COCO}: contains 82,783 training, 40,504 validation image-text pairs which belong to 91 categories. We select 38,000 image-text pairs that belong to the 20 most frequent concepts. The text for each point is represented as a 2912-dimensional bag-of-words vector; {\bf WKG Game-Hub}: consists of 13,750 articles collected from the Game-Hub of `` Strike of Kings'' with 1744 concept labels. We select 11,000 image-text pairs that belong to the 54 most frequent concepts. Each article contains several images and content paragraphs, and the text for each point is represented as a 300-dimensional w2v vector.

We run each compared method 30 times for all datasets, and then randomly select $70\%$ for training and the remaining are for test. For all the training examples, we randomly choose $30\%$ as the labeled data, and the other $70\%$ as unlabeled ones as~\cite{ZhangLLG18}. For the 4 benchmark datasets, each image is divided into 10 regions using~\cite{Girshick15} as image bag, while the corresponding text tags are also separated into several independent tags as text bag. For the WKG Game-Hub dataset, each article is denoted as an image bag and a content bag. The deep network for image encoder is implemented the same as Resnet-18~\cite{He2015}. We run the following experiments with the implementation of an environment on NVIDIA K80 GPUs server, and our model can be trained around 290 images per second with a single K80 GPGPU. In the training phase, the parameters $\lambda_1$ is selected by 5-fold cross validation from $\{10^{-5},10^{-4}, \cdots,10^4, 10^5\}$ with further splitting on only the training datasets, i.e., there is no overlap between the test set and the validation set for parameter picking up. Empirically, when the variation between the objective values of Eq. \ref{eq:e8} is less than $10^{-6}$ in iteration, we treat M3DN or M3DNS converged.

\begin{figure*}[t]
\begin{center}
\begin{minipage}[h]{78mm}
\centering
\includegraphics[width=78mm ]{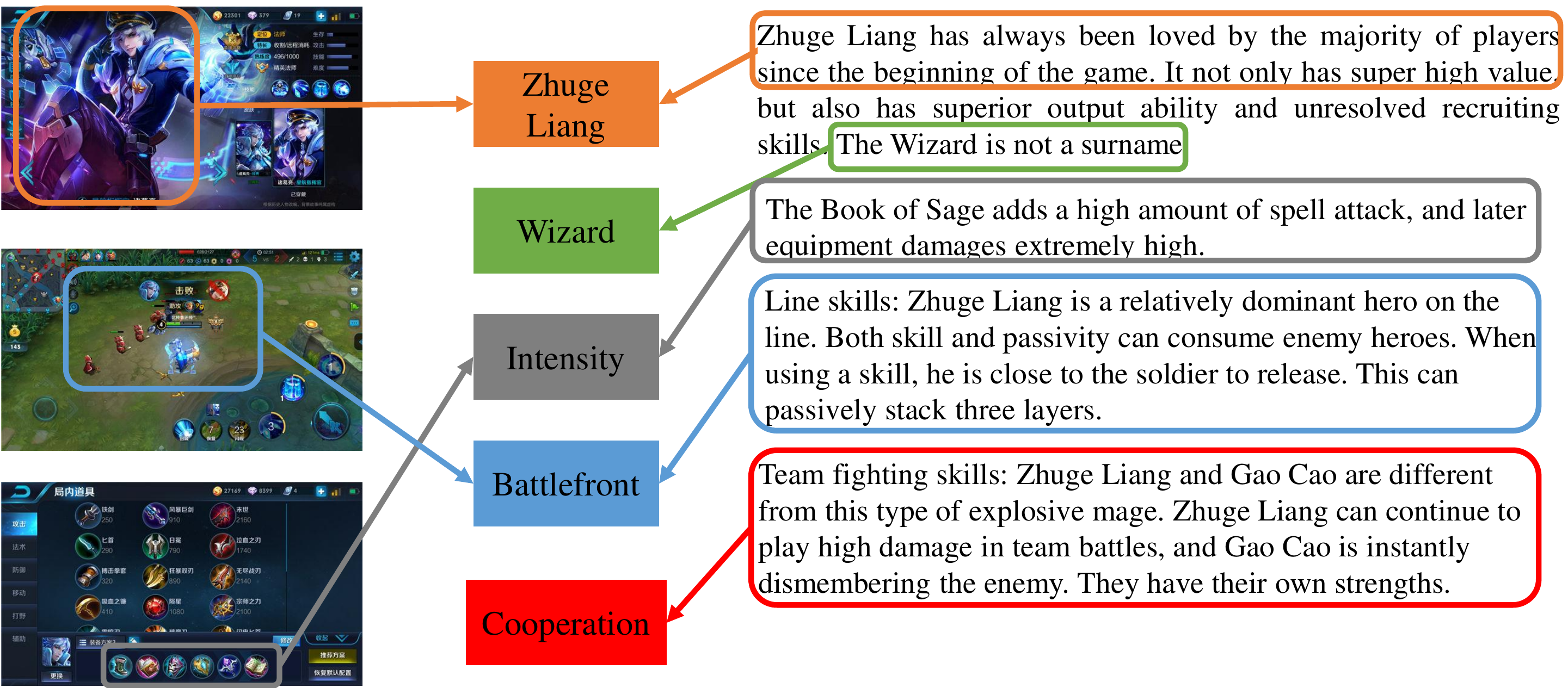}
\end{minipage}
\begin{minipage}[h]{78mm}
\centering
\includegraphics[width=78mm ]{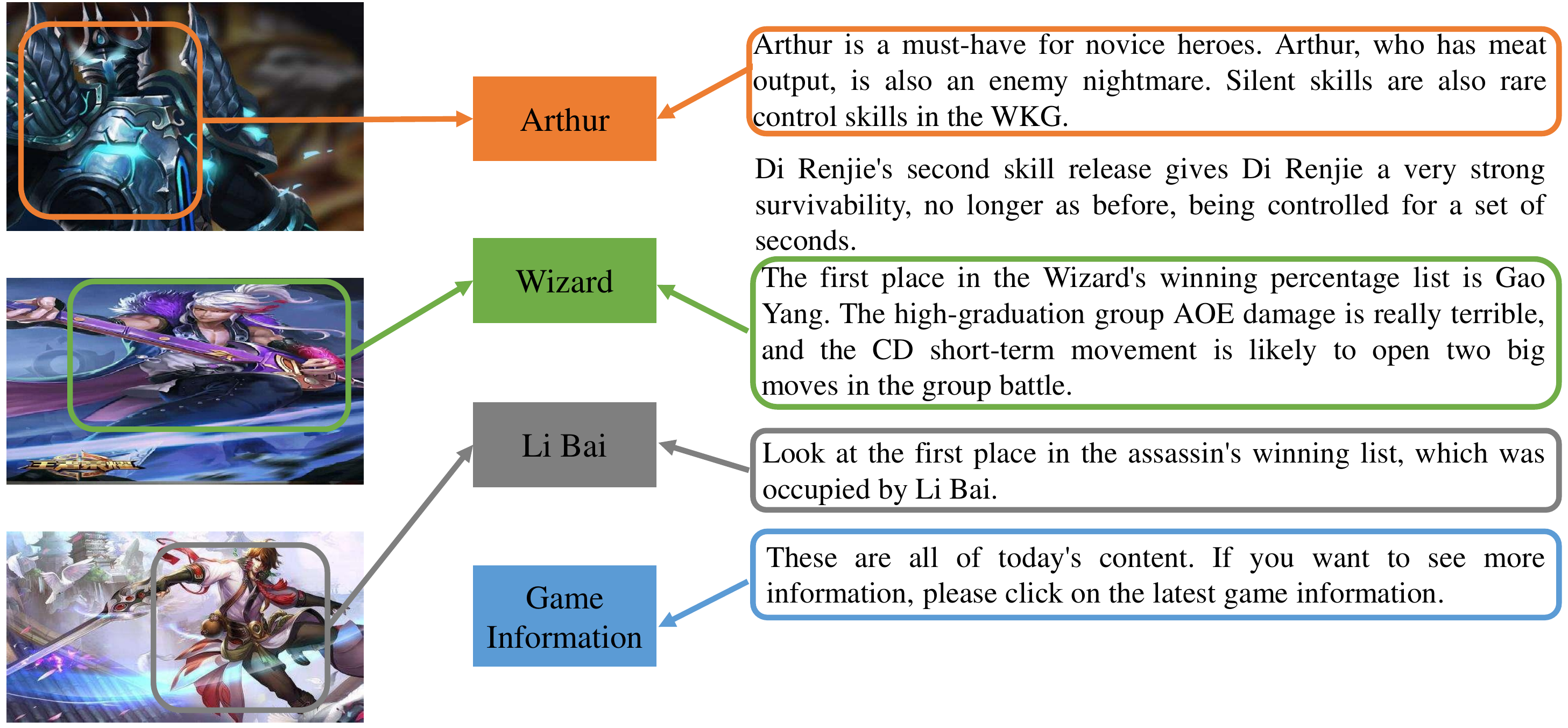}\\
\end{minipage}\\
\begin{minipage}[h]{78mm}
\centering
\includegraphics[width=78mm ]{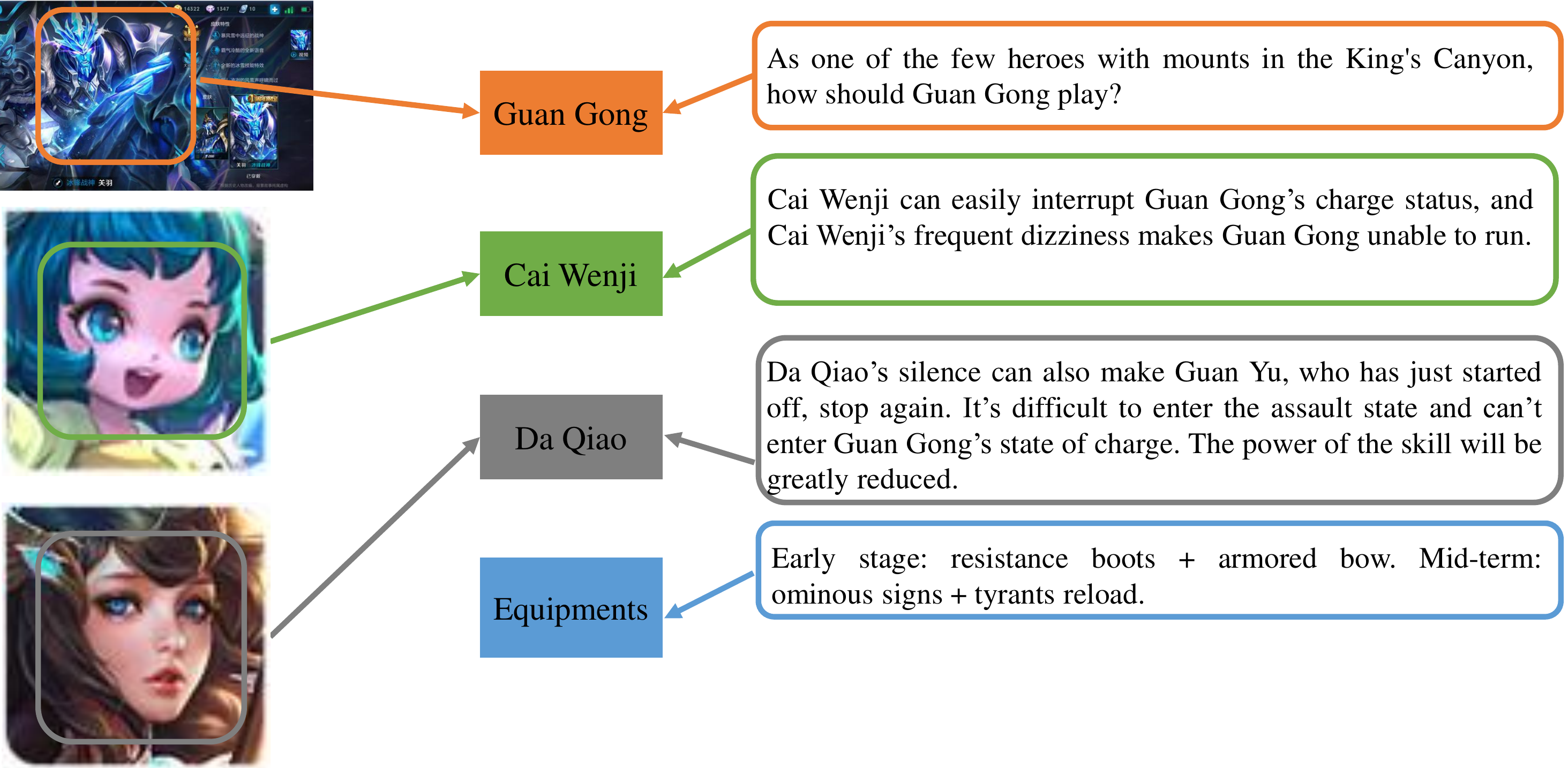}
\end{minipage}
\begin{minipage}[h]{78mm}
\centering
\includegraphics[width=78mm ]{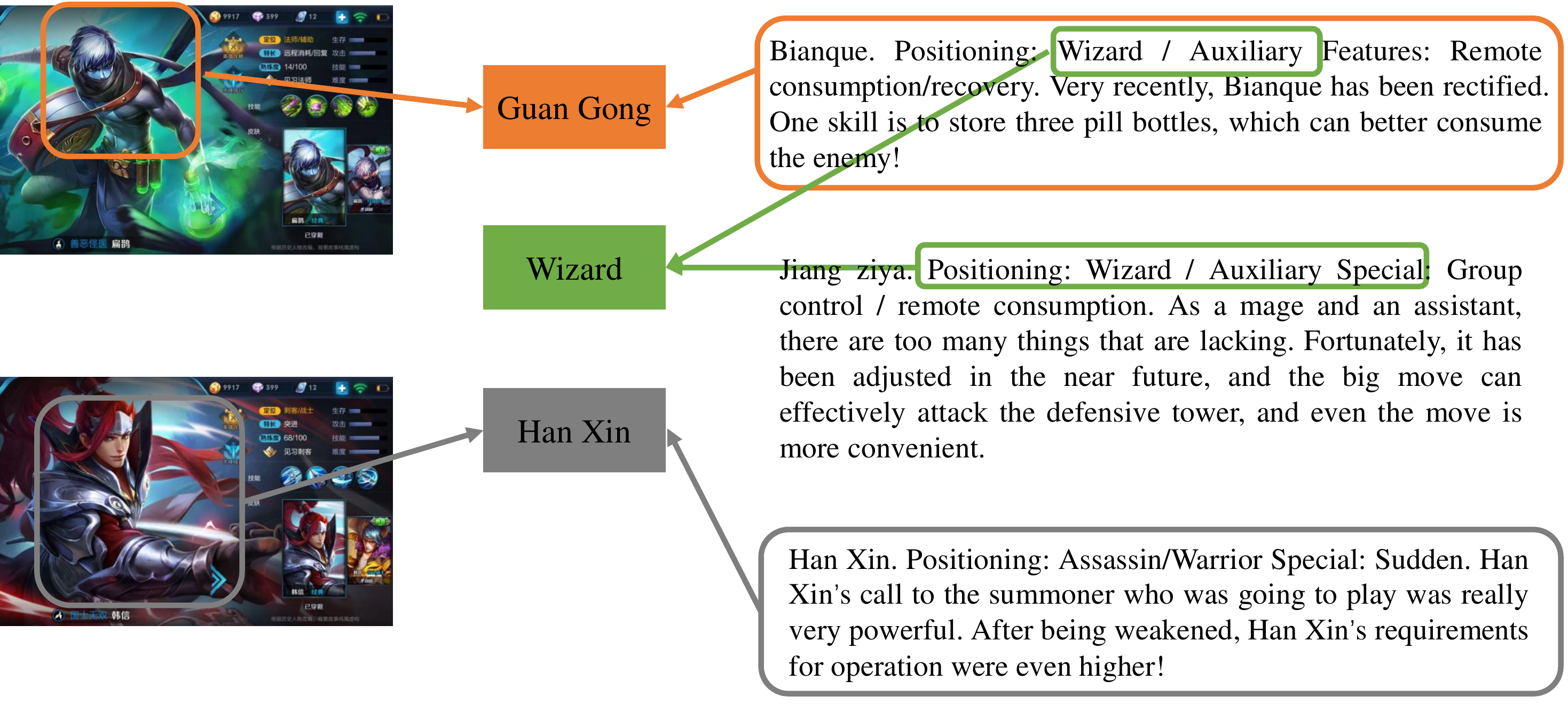}\\
\end{minipage}\\
\begin{minipage}[h]{78mm}
\centering
\includegraphics[width=78mm ]{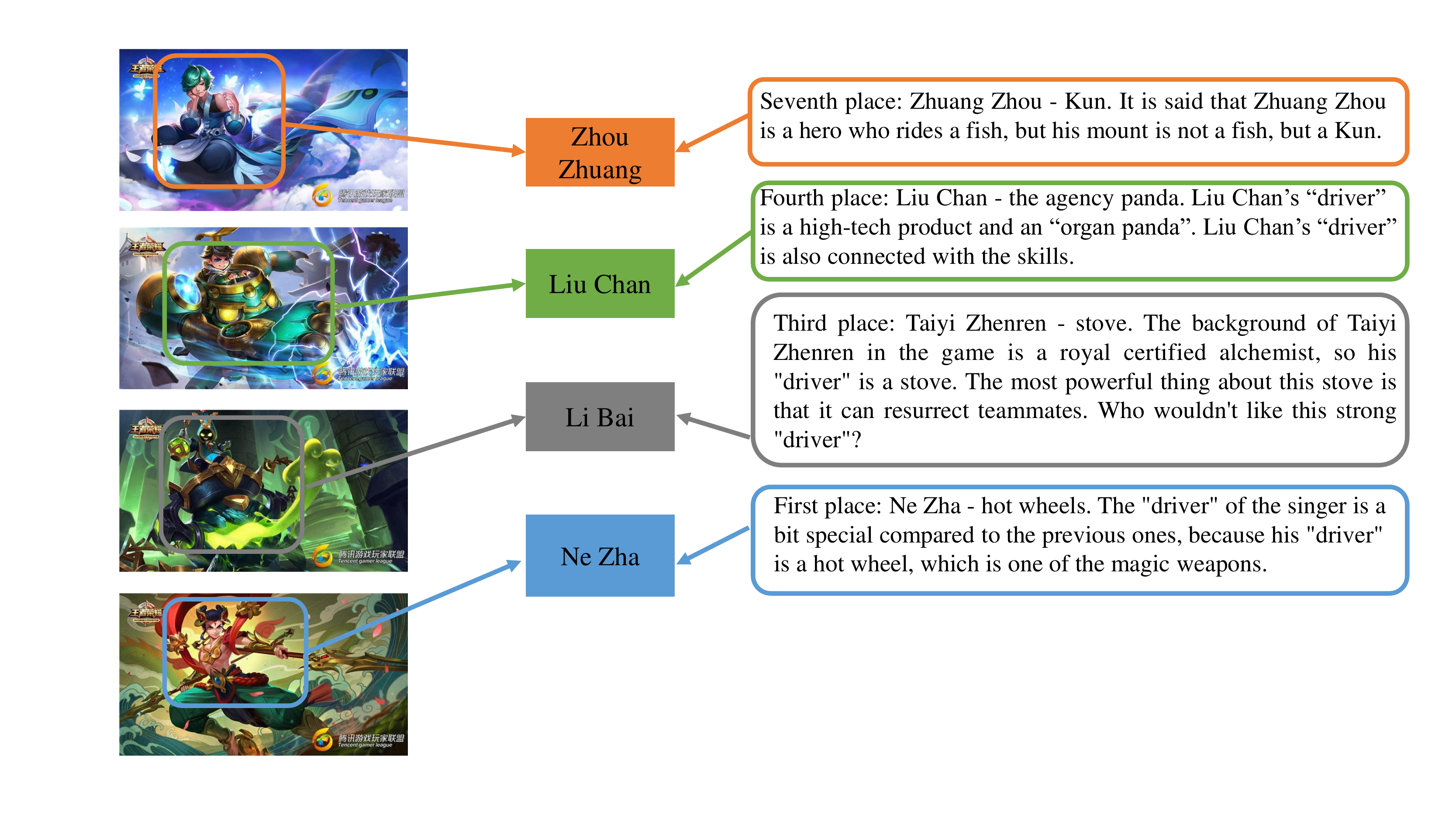}
\end{minipage}
\begin{minipage}[h]{78mm}
\centering
\includegraphics[width=78mm ]{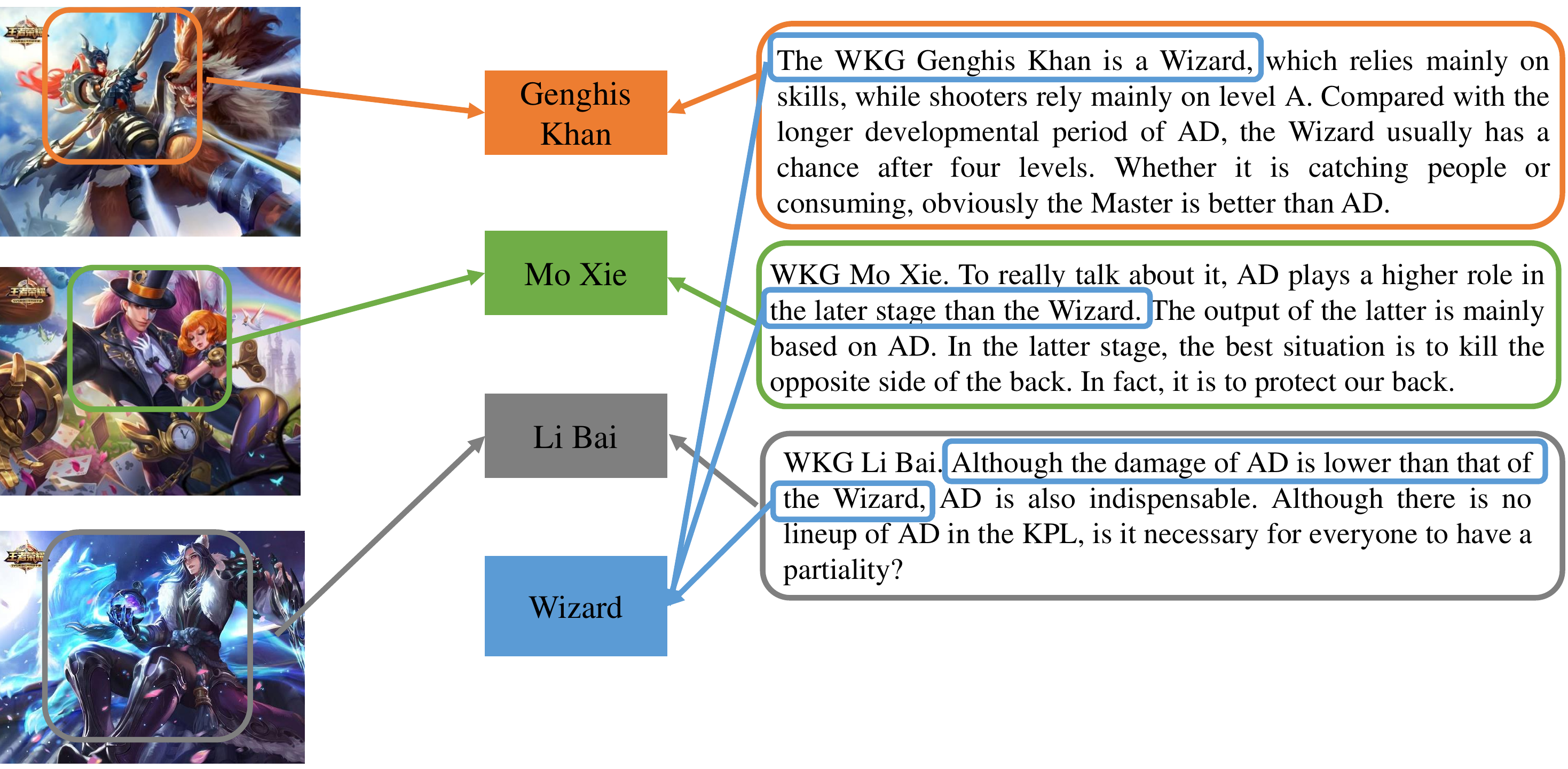}\\
\end{minipage}
\end{center}
\caption{Sample test complex articles predictions of the WKG Game-Hub. Left is the image bag, middle are label predictions, right is the context bag.}\label{fig:res1}
\end{figure*}

\subsection{Compared methods}
% MIMLmix~\cite{NguyenWLZ14}
In our experiments, first, we compare our methods with multi-modal multi-instance multi-label methods, i.e., M3LDA~\cite{NguyenZZ13}, MIMLmix~\cite{NguyenWLZ14}. Besides, M3DN can be degenerated into different settings, we also compare with multi-modal multi-label methods, i.e., CS3G~\cite{YeZLHJ16}; multi-instance multi-label methods, i.e., DeepMIML~\cite{FengZ17}, M3MIML~\cite{ZhangZ08}, MIMLfast~\cite{HuangGZ14}. Moreover, we compare our methods with multi-label methods, i.e.,  SLEEC~\cite{BhatiaJKVJ15}, Tram~\cite{KongNZ13}, ECC~\cite{ReadPHF11}, ML-KNN~\cite{ZhangZ07}, RankSVM~\cite{Joachims02}, ML-SVM~\cite{BoutellLSB04}. Specifically, for multi-modal multi-label methods, we calculate the average of all instances' representations as the bag-level feature representation. In the multi-instance multi-label methods, all modalities of a dataset are concatenated together as a single modal input. As to the multi-label learners, we first calculate bag-level feature representation for different modalities independently, then we concatenate all modalities together as a single modal input. As to the semi-supervised scenario, considering that existing M3 methods are supervised methods, we compare our methods with semi-supervised multi-modal multi-label methods, i.e., CS3G~\cite{YeZLHJ16}; and semi-supervised multi-label methods, i.e., Tram~\cite{KongNZ13}, COINS~\cite{ZhanZ17}, iMLU~\cite{WuZ13}.

%In detail, the compared methods are listed as:

%% FastXML~\cite{PrabhuV14}, LEML~\cite{Yu0KD14},
%
%\noindent{\bf M3LDA}: makes the topic decided by the visual information and the topic decided by the textual information to be consistent, leading to the label assignment;
%
%\noindent{\bf MIMLmix}: is a hierarchical Bayesian network, the instances are sampled from a mixture model where components are representations of labels in multiple modalities;
%
%\noindent{\bf CS3G}: handles types of interactions between multiple labels and utilizes the data from different modalities;
%
%\noindent{\bf DeepMIML}: exploits deep neural network to
%generate instance representation for MIML;
%
%\noindent{\bf M3MIML}: learns from multi-instance multi-label examples by maximum margin strategy;
%
%\noindent{\bf MIMLfast}: is a fast multi-instance multi-label method;
%
%\noindent{\bf SLEEC}: learns a small ensemble of local distance preserving embeddings, which can accurately predict infrequently occurring labels;
%
%\noindent{\bf Tram}: is a transductive multi-label classification algorithm via label set propagation;
%
%\noindent{\bf ECC}: state-of-the-art supervised ensemble multi-label method;
%
%\noindent{\bf ML-KNN}: is a kNN style multi-label classification algorithm which often outperforms other existing multi-label algorithms;
%
%\noindent{\bf RankSVM}: learns a ranking function for multi-label classification;
%
%\noindent{\bf ML-SVM}: proposes a new training strategy, i.e., cross-training, to build multi-label classifiers.

\begin{table*}[t]{\small
\centering
\caption{Semi-supervised comparison results (mean $\pm$ std.) of M3DNS with compared methods on 4 benchmark datasets. 6 commonly used criteria are evaluated. The best performance for each criterion is bolded. $\uparrow/\downarrow$ indicates the larger/smaller the better of the criterion.}
\label{tab:add}
\begin{tabular*}{1\textwidth}{@{\extracolsep{\fill}}@{}l|@{}c@{}|@{}c@{\;}|@{}@{}c@{}@{\;}|c|c|c|c|c@{}}
\toprule
\multirow{2}{*}{Methods} & \multicolumn{4}{c|}{Coverage $\downarrow$} & \multicolumn{4}{c}{Macro AUC $\uparrow$}\\
\cmidrule(l){2-9}
& FLICKR25K & IAPR TC-12 & MS-CoCo & NUS-WIDE & FLICKR25K & IAPRTC-12 & MS-CoCo & NUS-WIDE \\
\midrule
CS3G  &10.346$\pm$.227 &7.545$\pm$.056 &6.968$\pm$.060 &9.819$\pm$.931 & .844$\pm$.006 & .798$\pm$.002 &.699$\pm$.006 &.662$\pm$.077\\ \midrule
Tram  &6.857$\pm$.645 &5.793$\pm$.359 &55.059$\pm$1.888 &9.359$\pm$.223 & .827$\pm$.001 & .805$\pm$.001 &\bf .891$\pm$.001 &.890$\pm$.045\\
COINS  &22.940$\pm$5.082 &20.598$\pm$4.513 &25.839$\pm$10.629 &20.126$\pm$4.072 & .891$\pm$.004 & .863$\pm$.006 &.814$\pm$.014 &.873$\pm$.017\\
iMLU  &23.411$\pm$1.160 &23.401$\pm$8.939 &26.462$\pm$5.548 &21.030$\pm$4.844 & .880$\pm$.009 & .835$\pm$.003 &.812$\pm$.004 &.835$\pm$.048\\\midrule
M3DNS &\bf 3.947$\pm$.307 &\bf 4.214$\pm$.202 &\bf 6.119$\pm$.262 &\bf 2.764$\pm$.071 &\bf.892$\pm$.004 &\bf .876$\pm$.003  &.838$\pm$.003  &\bf .898$\pm$.008 \\
\end{tabular*}
\begin{tabular*}{1\textwidth}{@{\extracolsep{\fill}}@{}l|c|c|c|c|c|c|c|c@{}}
\toprule
\multirow{2}{*}{Methods} & \multicolumn{4}{c|}{Ranking Loss $\downarrow$} & \multicolumn{4}{c}{Example AUC $\uparrow$}\\
\cmidrule(l){2-9}
& FLICKR25K & IAPR TC-12 & MS-CoCo & NUS-WIDE & FLICKR25K & IAPRTC-12 & MS-CoCo & NUS-WIDE \\
\midrule
CS3G  &.109$\pm$.003 &.120$\pm$.001 &.168$\pm$.001 &.196$\pm$.070 & .890$\pm$.003 & .879$\pm$.001 &.831$\pm$.001 &.803$\pm$.070\\ \midrule
Tram  &.108$\pm$.002 &.119$\pm$.001 &.183$\pm$.001 &.183$\pm$.076 & .893$\pm$.002 &\bf .880$\pm$.001 &.816$\pm$.001 &.816$\pm$.076\\
COINS  &.150$\pm$.009 &.171$\pm$.002 &.305$\pm$.008 &.297$\pm$.028 & .849$\pm$.009 & .828$\pm$.002 &.694$\pm$.008 &.702$\pm$.028\\
iMLU  &.167$\pm$.007 &.242$\pm$.014 &.344$\pm$.013 &.346$\pm$.015 & .832$\pm$.007 & .757$\pm$.014 &.655$\pm$.013 &.653$\pm$.015\\\midrule
M3DNS &\bf .108$\pm$.001 &\bf .142$\pm$.002 &\bf .112$\pm$.003 &\bf .119$\pm$.003 &\bf .899$\pm$.004 & .858$\pm$.005  &\bf .898$\pm$.008  &\bf .881$\pm$.006 \\
\end{tabular*}
\begin{tabular*}{1\textwidth}{@{\extracolsep{\fill}}@{}l|c|c|c|c|c|c|c|c@{}}
\toprule
\multirow{2}{*}{Methods} & \multicolumn{4}{c|}{Average Precision $\uparrow$} & \multicolumn{4}{c}{Micro AUC $\uparrow$}\\
\cmidrule(l){2-9}
& FLICKR25K & IAPR TC-12 & MS-CoCo & NUS-WIDE & FLICKR25K & IAPRTC-12 & MS-CoCo & NUS-WIDE \\
\midrule
CS3G  &.671$\pm$.003 &\bf.678$\pm$.001 &.661$\pm$.003 &.586$\pm$.083 & .860$\pm$.007 & .820$\pm$.002 &.769$\pm$.003 &.724$\pm$.084\\ \midrule
Tram  &.670$\pm$.006 &.507$\pm$.004 &.348$\pm$.003 &.318$\pm$.091 &\bf .910$\pm$.001 & .859$\pm$.001 &.874$\pm$.001 &.868$\pm$.057\\
COINS  &.570$\pm$.007 &.419$\pm$.007 &.258$\pm$.033 &.216$\pm$.016 & .884$\pm$.007 & .852$\pm$.003 &.788$\pm$.018 &.856$\pm$.025\\
iMLU  &.538$\pm$.015 &.325$\pm$.016 &.220$\pm$.043 &.187$\pm$.015 & .860$\pm$.015 & .793$\pm$.007 &.760$\pm$.013 &.798$\pm$.078 \\\midrule
M3DNS &\bf .698$\pm$.002 & .637$\pm$.007 &\bf .691$\pm$.004 &\bf .634$\pm$.003 &.858$\pm$.003 &\bf .863$\pm$.004  &\bf .877$\pm$.006  &\bf .878$\pm$.005 \\
\bottomrule
\end{tabular*}}
\end{table*}

\begin{table*}[t]\centering{
\caption{Semi-supervised comparison results (mean $\pm$ std.) of M3DNS with compared methods on WKG Game-Hub dataset. 6 commonly used criteria are evaluated. The best performance for each criterion is bolded. $\uparrow/\downarrow$ indicates the larger/smaller, the better of the criterion.}
  \label{tab:add1}
  \begin{tabular*}{1\textwidth}{@{\extracolsep{\fill}}@{}l|c|c|c|c|c|c}
    \addlinespace
    \toprule
    Methods & {Coverage $\downarrow$  ($\times 10^3$)} & Macro AUC $\uparrow$  & Ranking Loss $\downarrow$ & Example AUC $\uparrow$ & Average Precision $\uparrow$ &Micro AUC $\uparrow$\\
    \midrule
    CS3G  &.326$\pm$.002 &.683$\pm$.021 &.187$\pm$.014 &.812$\pm$.014 & .404$\pm$.057 & .728$\pm$.026 \\ \midrule
    Tram  &1.731$\pm$.083 &.854$\pm$.031  &.190$\pm$.024 &.809$\pm$.024 & .245$\pm$.046 & .852$\pm$.024 \\
    COINS  &.186$\pm$.021 &.782$\pm$.087 &.252$\pm$.029 &.747$\pm$.029 & .195$\pm$.037 & .783$\pm$.072 \\
    iMLU  &.225$\pm$.027 &.786$\pm$.070 &.288$\pm$.033 &.711$\pm$.030 & .169$\pm$.026 & .763$\pm$.010 \\\midrule
    M3DNS &\bf .149$\pm$.002 &\bf .933$\pm$.001 &\bf .180$\pm$.009 &\bf .828$\pm$.003 &\bf .409$\pm$.001 &\bf .880$\pm$.001\\
    \bottomrule
    \end{tabular*}}
\end{table*}

\begin{table*}[t]{\small
\centering
\caption{Ablation study results (mean $\pm$ std.) of M3DNS on 4 benchmark datasets. 6 commonly used criteria are evaluated. The best performance for each criterion is bolded. $\uparrow/\downarrow$ indicates the larger/smaller the better of the criterion.}
\label{tab:add3}
\begin{tabular*}{1\textwidth}{@{\extracolsep{\fill}}@{}l|c|@{}c|@{}@{}c@{}@{\;}|c|c|c|c|c@{}}
\toprule
\multirow{2}{*}{Methods} & \multicolumn{4}{c|}{Coverage $\downarrow$} & \multicolumn{4}{c}{Macro AUC $\uparrow$}\\
\cmidrule(l){2-9}
& FLICKR25K & IAPR TC-12 & MS-CoCo & NUS-WIDE & FLICKR25K & IAPRTC-12 & MS-CoCo & NUS-WIDE \\
\midrule
M3DNS-F  &8.678$\pm$.002 &6.875$\pm$.010 &9.280$\pm$.003 &11.042$\pm$.009 &\bf.896$\pm$.000 & .868$\pm$.000 &.829$\pm$.002 &.858$\pm$.001\\
M3DNS-M  &8.889$\pm$.010 &6.964$\pm$.003 &9.764$\pm$.001 &11.043$\pm$.005 & .885$\pm$.001 & .862$\pm$.000 &.757$\pm$.001 &.843$\pm$.000\\
M3DNS-MP  &4.039$\pm$.021 &5.047$\pm$.038 &.8.708$\pm$.028 &3.230$\pm$.003 & .874$\pm$.000 & .860$\pm$.000 &.779$\pm$.001 &.837$\pm$.001\\
M3DNS &\bf 3.947$\pm$.307 &\bf 4.214$\pm$.202 &\bf 6.119$\pm$.262 &\bf 2.764$\pm$.071 &.892$\pm$.004 &\bf .876$\pm$.003  &\bf .838$\pm$.003  &\bf .898$\pm$.008 \\
\end{tabular*}
\begin{tabular*}{1\textwidth}{@{\extracolsep{\fill}}@{}l|c|c|c|c|c|c|c|c@{}}
\toprule
\multirow{2}{*}{Methods} & \multicolumn{4}{c|}{Ranking Loss $\downarrow$} & \multicolumn{4}{c}{Example AUC $\uparrow$}\\
\cmidrule(l){2-9}
& FLICKR25K & IAPR TC-12 & MS-CoCo & NUS-WIDE & FLICKR25K & IAPRTC-12 & MS-CoCo & NUS-WIDE \\
\midrule
M3DNS-F  &\bf .074$\pm$.000 &.146$\pm$.000 &.134$\pm$.001 &.184$\pm$.000  &.825$\pm$.000 &.804$\pm$.000 &.866$\pm$.001  &.816$\pm$.000\\
M3DNS-M  &.109$\pm$.001 &.149$\pm$.000 &.150$\pm$.000 &.132$\pm$.000 & .783$\pm$.001 & .696$\pm$.000 &.686$\pm$.000 &.540$\pm$.001\\
M3DNS-MP  &.106$\pm$.000 & .145$\pm$.001 & .150$\pm$.001 &.190$\pm$.001  &.818$\pm$.000 & .790$\pm$.001 & .848$\pm$.000 &.810$\pm$.001\\
M3DNS & .108$\pm$.001 &\bf .142$\pm$.002 &\bf .112$\pm$.003 &\bf .119$\pm$.003 &\bf .899$\pm$.004 &\bf .858$\pm$.005  &\bf .898$\pm$.008  &\bf .881$\pm$.006 \\
\end{tabular*}
\begin{tabular*}{1\textwidth}{@{\extracolsep{\fill}}@{}l|c|c|c|c|c|c|c|c@{}}
\toprule
\multirow{2}{*}{Methods} & \multicolumn{4}{c|}{Average Precision $\uparrow$} & \multicolumn{4}{c}{Micro AUC $\uparrow$}\\
\cmidrule(l){2-9}
& FLICKR25K & IAPR TC-12 & MS-CoCo & NUS-WIDE & FLICKR25K & IAPRTC-12 & MS-CoCo & NUS-WIDE \\
\midrule
M3DNS-F  &.693$\pm$.000 &.592$\pm$.000 &\bf .693$\pm$.000 &.624$\pm$.000 &\bf .917$\pm$.000 & .863$\pm$.002 &.868$\pm$.003 &.877$\pm$.000\\
M3DNS-M  &.614$\pm$.002 &.588$\pm$.000 &.639$\pm$.001 &.610$\pm$.000 & .819$\pm$.001 & .790$\pm$.000 &.850$\pm$.003 &.814$\pm$.001\\
M3DNS-MP  &.681$\pm$.000 &.582$\pm$.001 &.684$\pm$.001 &.616$\pm$.001 & .809$\pm$.000 & .791$\pm$.000 &.846$\pm$.001 &.807$\pm$.002\\
M3DNS &\bf .698$\pm$.002 &\bf .637$\pm$.007 & .691$\pm$.004 &\bf .634$\pm$.003 & .858$\pm$.003 &\bf .863$\pm$.004  &\bf .877$\pm$.006  &\bf .878$\pm$.005 \\
\bottomrule
\end{tabular*}}
\end{table*}

\begin{table*}[t]\centering{
\caption{Ablation study results (mean $\pm$ std.) of M3DNS on WKG Game-Hub dataset. 6 commonly used criteria are evaluated. The best performance for each criterion is bolded. $\uparrow/\downarrow$ indicates the larger/smaller, the better of the criterion.}
  \label{tab:add4}
  \begin{tabular*}{1\textwidth}{@{\extracolsep{\fill}}@{}l|c|c|c|c|c|c}
    \addlinespace
    \toprule
    Methods & {Coverage $\downarrow$  ($\times 10^3$)} & Macro AUC $\uparrow$  & Ranking Loss $\downarrow$ & Example AUC $\uparrow$ & Average Precision $\uparrow$ &Micro AUC $\uparrow$\\
    \midrule
    M3DNS-F  &.279$\pm$.003 &.821$\pm$.000 &.183$\pm$.001 &.822$\pm$.000 & .345$\pm$.000 & .872$\pm$.000 \\
    M3DNS-M  &.287$\pm$.041 &.840$\pm$.000 &.182$\pm$.001 &.823$\pm$.000 & .379$\pm$.001 & .870$\pm$.002 \\
    M3DNS-MP  &.286$\pm$.008 &.818$\pm$.000 &.190$\pm$.001 &.817$\pm$.001 & .333$\pm$.000 & .869$\pm$.002 \\
    M3DNS &\bf .149$\pm$.002 &\bf .933$\pm$.001 &\bf .180$\pm$.009 &\bf .828$\pm$.003 &\bf .409$\pm$.001 &\bf .880$\pm$.001\\
    \bottomrule
    \end{tabular*}}
\end{table*}

\begin{table*}[t]{\small
\centering
\caption{Missing modal comparison results (mean $\pm$ std.) of M3DNS on 4 benchmark datasets. 6 commonly used criteria are evaluated. The best performance for each criterion is bolded. $\uparrow/\downarrow$ indicates the larger/smaller the better of the criterion.}
\label{tab:add5}
\begin{tabular*}{1\textwidth}{@{\extracolsep{\fill}}@{}l|c|c|c@{\;}|c|c|c|c|c@{}}
\toprule
\multirow{2}{*}{Methods} & \multicolumn{4}{c|}{Coverage $\downarrow$} & \multicolumn{4}{c}{Macro AUC $\uparrow$}\\
\cmidrule(l){2-9}
& FLICKR25K & IAPR TC-12 & MS-CoCo & NUS-WIDE & FLICKR25K & IAPRTC-12 & MS-CoCo & NUS-WIDE \\
\midrule
$0\%$ &\bf 3.947$\pm$.307 &\bf 4.214$\pm$.202 &\bf 6.119$\pm$.262 &\bf 2.764$\pm$.071 &\bf .892$\pm$.004 &\bf .876$\pm$.003  &\bf .838$\pm$.003  &\bf .898$\pm$.008 \\
$10\%$  &4.012$\pm$.013 &5.017$\pm$.015 &6.443$\pm$.002 &2.815$\pm$.018 & .891$\pm$.000 & .858$\pm$.001 &.822$\pm$.000 &.865$\pm$.001\\
$30\%$  &4.033$\pm$.009 &5.604$\pm$.013 &6.324$\pm$.007 &2.834$\pm$.010 & .888$\pm$.001 & .870$\pm$.001 &.817$\pm$.001 &.866$\pm$.000\\
$50\%$  &4.080$\pm$.003 &5.862$\pm$.000 &6.496$\pm$.004 &3.381$\pm$.002 & .887$\pm$.000 & .862$\pm$.004 &.812$\pm$.000 &.834$\pm$.001\\
$70\%$  &4.180$\pm$.021 &5.840$\pm$.002 &6.378$\pm$.005 &3.213$\pm$.001 & .880$\pm$.000 & .861$\pm$.000 &.806$\pm$.001 &.846$\pm$.000\\
$90\%$  &4.485$\pm$.004 &5.897$\pm$.001 &6.816$\pm$.017 &3.615$\pm$.004 & .869$\pm$.000 & .856$\pm$.000 &.781$\pm$.000 &.820$\pm$.001\\
\end{tabular*}
\begin{tabular*}{1\textwidth}{@{\extracolsep{\fill}}@{}l|c|c|c|c|c|c|c|c@{}}
\toprule
\multirow{2}{*}{Methods} & \multicolumn{4}{c|}{Ranking Loss $\downarrow$} & \multicolumn{4}{c}{Example AUC $\uparrow$}\\
\cmidrule(l){2-9}
& FLICKR25K & IAPR TC-12 & MS-CoCo & NUS-WIDE & FLICKR25K & IAPRTC-12 & MS-CoCo & NUS-WIDE \\
\midrule
$0\%$ &\bf .108$\pm$.001 &\bf .142$\pm$.002 &\bf .112$\pm$.003 &\bf .119$\pm$.003 &\bf .899$\pm$.004 &\bf .858$\pm$.005  &\bf .898$\pm$.008  &\bf .881$\pm$.006 \\
$10\%$  &.178$\pm$.000 &.159$\pm$.000 &.140$\pm$.000 &.178$\pm$.000 & .892$\pm$.000 & .840$\pm$.000 &.859$\pm$.000 &.871$\pm$.000\\
$30\%$  &.180$\pm$.000 &.150$\pm$.001 &.138$\pm$.000 &.178$\pm$.000 & .879$\pm$.000 & .849$\pm$.000 &.861$\pm$.001 &.871$\pm$.000\\
$50\%$  &.181$\pm$.000 &.157$\pm$.000 &.143$\pm$.000 &.192$\pm$.000 & .878$\pm$.001 & .842$\pm$.000 &.856$\pm$.000 &.857$\pm$.000\\
$70\%$  &.185$\pm$.001 &.155$\pm$.000 &.139$\pm$.000 &.187$\pm$.001 & .874$\pm$.001 & .844$\pm$.000 &.854$\pm$.000 &.862$\pm$.004\\
$90\%$  &.190$\pm$.002 &.159$\pm$.001 &.156$\pm$.000 &.199$\pm$.000 & .869$\pm$.000 & .839$\pm$.001 &.843$\pm$.000 &.850$\pm$.000\\
\end{tabular*}
\begin{tabular*}{1\textwidth}{@{\extracolsep{\fill}}@{}l|c|c|c|c|c|c|c|c@{}}
\toprule
\multirow{2}{*}{Methods} & \multicolumn{4}{c|}{Average Precision $\uparrow$} & \multicolumn{4}{c}{Micro AUC $\uparrow$}\\
\cmidrule(l){2-9}
& FLICKR25K & IAPR TC-12 & MS-CoCo & NUS-WIDE & FLICKR25K & IAPRTC-12 & MS-CoCo & NUS-WIDE \\
\midrule
$0\%$   &\bf .698$\pm$.002 &\bf .637$\pm$.007 &\bf .691$\pm$.004 &\bf .634$\pm$.003 & .858$\pm$.003 &\bf .863$\pm$.004  &\bf .877$\pm$.006  &\bf .878$\pm$.005 \\
$10\%$  &.689$\pm$.000 &.631$\pm$.000 &.684$\pm$.000 &.631$\pm$.000 & .817$\pm$.000 & .845$\pm$.000 &.860$\pm$.000 &.870$\pm$.000\\
$30\%$  &.678$\pm$.000 &.635$\pm$.000 &.686$\pm$.002 &.631$\pm$.000 & .812$\pm$.000 & .855$\pm$.002 &.862$\pm$.001 &.869$\pm$.000\\
$50\%$  &.678$\pm$.000 &.628$\pm$.000 &.679$\pm$.001 &.598$\pm$.000 & .815$\pm$.000 & .849$\pm$.000 &.857$\pm$.000 &.853$\pm$.000\\
$70\%$  &.666$\pm$.001 &.629$\pm$.000 &.680$\pm$.000 &.593$\pm$.000 & .808$\pm$.001 & .848$\pm$.000 &.862$\pm$.000 &.858$\pm$.000\\
$90\%$  &.659$\pm$.000 &.610$\pm$.000 &.663$\pm$.001 &.590$\pm$.000 & .802$\pm$.000 & .846$\pm$.000 &.842$\pm$.000 &.846$\pm$.000\\
\bottomrule
\end{tabular*}}
\end{table*}

\begin{table*}[t]\centering{
\caption{Missing modal comparison results (mean $\pm$ std.) of M3DNS on WKG Game-Hub dataset. 6 commonly used criteria are evaluated. The best performance for each criterion is bolded. $\uparrow/\downarrow$ indicates the larger/smaller, the better of the criterion.}
  \label{tab:add6}
  \begin{tabular*}{1\textwidth}{@{\extracolsep{\fill}}@{}l|c|c|c|c|c|c}
    \addlinespace
    \toprule
    Methods & {Coverage $\downarrow$  ($\times 10^3$)} & Macro AUC $\uparrow$  & Ranking Loss $\downarrow$ & Example AUC $\uparrow$ & Average Precision $\uparrow$ &Micro AUC $\uparrow$\\
    \midrule
    $0\%$ &\bf .149$\pm$.002 &\bf .933$\pm$.001 &\bf .180$\pm$.009 &\bf .828$\pm$.003 &\bf .409$\pm$.001 &\bf .880$\pm$.001\\
    $10\%$  &.264$\pm$.007 &.844$\pm$.000 &.183$\pm$.000 &.776$\pm$.000 & .379$\pm$.000 & .877$\pm$.000 \\
    $30\%$  &.273$\pm$.003 &.830$\pm$.000 &.191$\pm$.000 &.768$\pm$.001 & .363$\pm$.000 & .868$\pm$.000 \\
    $50\%$  &.276$\pm$.013 &.825$\pm$.000 &.193$\pm$.000 &.766$\pm$.000 & .350$\pm$.000 & .866$\pm$.000 \\
    $70\%$  &.284$\pm$.002 &.812$\pm$.000 &.201$\pm$.000 &.758$\pm$.000 & .336$\pm$.000 & .859$\pm$.000 \\
    $90\%$  &.299$\pm$.008 &.802$\pm$.000 &.207$\pm$.000 &.752$\pm$.000 & .329$\pm$.001 & .848$\pm$.000 \\
    \bottomrule
    \end{tabular*}}
\end{table*}

\subsection{Benchmark Comparisons}
M3DN is compared with other methods on 4 benchmark datasets to demonstrate the abilities. Results of compared methods and M3DN/M3DNS on 6 commonly used criteria are listed in Tab. ~\ref{tab:tab11}. The best performance for each criterion is bolded. $\uparrow/\downarrow$ indicates that the larger/smaller, the better of the criterion. From the results, it is obvious that our M3DN/M3DNS approaches can achieve the best or second performance on most datasets with different performance measures. Therefore the M3DN/M3DNS approach are highly competitive multi-modal multi-label learning methods.

\subsection{Complex Article Classification}
In this subsection, M3DN approach is tested on the real-world complex article classification problem, i.e.,  WKG Game-Hub dataset. There are 13,570 articles in collection, with image and text modalities to promote classification. Specifically, each article contains variable number of images and text paragraphs. Thus, each article can be divided into both image bag and text bag. Comparison results (independent modalities and overall) against compared methods are listed in Tab. ~\ref{tab:tab2}, where notation ``N/A'' means the method cannot give a result in 60 hours. We use the same 6 measurement criteria as in previous subsection, i.e., Coverage, Ranking Loss, Average Precision, Macro AUC, example AUC and Micro AUC. It is notable that multi-label methods concatenate all of the modal features, which have no independent modal classification performance. The results show that on both of the independent modalities and overall prediction, our M3DN and M3DNS approaches can get the best results over all criteria. The statistics validates the effectiveness of our method when solving the complex article classification problem.

\subsection{Label Correlations Exploration}
Since M3DN can learn label correlation explicitly, in this subsection, we examine effectiveness of M3DN in label correlations exploration. Due to page limitation, the exploration is conducted on the real-world dataset WKG Game-Hug. We randomly sampled 27 labels, with the learned ground metric shown in Figure ~\ref{fig:res2}, and scaled the original value in cost matrix into $[-1,1]$. Red color indicates a positive correlation, and blue indicates a negative correlation. We can see that the learned pairwise cost accords with intuitions. Taking a few examples, the cost between Overwatcha and Tencent indicates a very small correlation, and this is reasonable as the game Overwatch has no correlation with Tencent. While the cost between (Zhuge Liang, Wizard) indicates a very strong correlation, since Zhuge Liang belongs to the wizard role in the game.

\subsection{Empirical Investigation on Convergence}
To investigate the convergence of M3DN iterations empirically, we record the objective function value, i.e., the value of Eq. \ref{eq:e2} and the different criteria of classification performance of M3DN/M3DNS in each epoch. Due to page limits, results on WKG Game-Hug dataset are plotted in Fig. \ref{fig:convergence}. It clearly reveals that the objective function value decreases as the iterations increase, and all of the classification performance is stable after several iterations in Fig. \ref{fig:convergence}. Moreover, these additional experiment results indicate that our M3DN/M3DNS can converge fast, i.e., M3DN converges after 10 epoches.

\subsection{Empirical Illustrative Examples}
Figure~\ref{fig:res1} shows 6 illustrative examples of the classification results on WKG Game-Hub dataset. Qualitatively, illustration of the predictions clearly discovers the modal-instance-label relation on the test set. E.g., the first example shows that the article has separated three images and four content paragraphs. We can predict the Zhuge liang, battlefront labels from both the images and contents, and acquire the master, cooperation labels form the context.

\section{Conclusion}
This paper focuses on the issues of complex objects classification with semi-supervised M3 information, and extends our preliminary research~\cite{YangWZL018}. Complex objects, i.e., the articles, the videos, etc, can always be represented by multi-modal multi-instance information, with multiple labels. However, we usually only have bag-level consistency among different modalities. Therefore, Multi-modal Multi-instance Multi-label (M3) learning provides a framework for handling such task. Meanwhile, previous M3 methods rarely consider label correlation and unlabeled data. In this paper, we propose a novel Multi-modal Multi-instance Multi-label Deep Network (M3DN) framework, and exploit label correlation based on the Optimal Transport (OT) theory. Moreover, considering unlabel information, M3DNS utilizes the instance-label and bag-level unlabel information for more excellent performance. Experiments on the real world benchmark datasets and special complex article dataset WKG Game-Hub validate effectiveness of the proposed methods. Meanwhile, how to extend to multiple modalities is an interesting future work.

\appendices
\section{Semi-Supervised Classification}
M3DNS takes unlabeled instances into consideration, i.e., using auto-encoder for single modal network, and consistency among different modalities for joint predictions. Thus, in this section, we provide empirical investigations and performance comparisons of M3DNS with several state-of-the-art semi-supervised methods. The introduction to data configuration and comparison methods are in Section 4.1, 4.2. The results are recorded in Table \ref{tab:add} and Table \ref{tab:add1}. The results indicate that M3DNS approach can achieve the best or second performance on most datasets with different performance measures, thus M3DNS can make better use of unlabeled data.

\section{Ablation Study}
In order to explore the impact of different operators in the network structure, we conduct more experiments. In detail, 1) in order to verify different pooling methods to get bag-level prediction, we compare max pooling with mean pooling, denoted as M3DNS-M with mean pooling; 2) based on the better bag-level pooling method, we compare average prediction with max prediction to evaluate different ensemble methods for final predictions, denoted as M3DNS-MP with max operator; 3) based on the better pooling method and prediction operator, we fix the ground metric as the initial value without any change to explore the advantage of learning ground metric, denoted as M3DNS-F. The results are recorded in Table \ref{tab:add3} and Table \ref{tab:add4}. It is notable that M3DNS is with max pooling, mean prediction operator. The results reveal that max pooling are always better than the mean pooling in getting bag-level prediction. This is because there are often only a few positive examples in the bag that can represent the prediction of this bag, yet mean pooling will bring a lot of noise on the contrast. This phenomenon is also consistent with the assumption of multi-instance learning. Furthermore, the results reveal that mean prediction operator is always better than the max operator, which is also according with the ensemble learning methods. An interesting thing is that, though M3DNS is better than M3DNS-F on most datasets, it is worse on one dataset, i.e., FLICKR25K. This result shows that learning ground metric is not definitely effective. Considering the noise data, it may affect the learning of ground metric. Thus, how to modify the learning process or design a suitable initialization method could be an interesting future work.

\section{Comparison with Missing Modality}
Specifically, in order to explore the impact of modal missing scenario, we conduct more experiments. Following~\cite{LiJZ14}, in each split, we randomly select $10\%$ to $90\%$ of examples, with $20\%$ as interval, for homogeneous examples with complete modality. And the remaining are incomplete instances. The results are recorded in Table \ref{tab:add5} and Table \ref{tab:add6}. It shows that M3DNS achieves competitive results when comparing the results in Table \ref{tab:tab11}, \ref{tab:tab2},  \ref{tab:add3} and \ref{tab:add4} with missing modalities, and the performance of M3DNS increases faster than compared methods as incomplete ratio decreases.

\ifCLASSOPTIONcompsoc
%\else
%  % regular IEEE prefers the singular form
  \section*{Acknowledgment}
\fi
This research was supported by National Key R$\&$D Program of China (2018YFB1004300),  NSFC (61773198, 61632004, 61751306), NSFC-NRF Joint Research Project under Grant 61861146001, and Collaborative Innovation Center of Novel Software Technology and Industrialization, Postgraduate Research $\&$ Practice Innovation Program of Jiangsu province (KYCX18-0045).

\ifCLASSOPTIONcaptionsoff
  \newpage
\fi

\bibliographystyle{IEEEtranN}{\footnotesize
\bibliography{M3DN}}



% \end{thebibliography}
\vspace{-1.2cm}
\begin{IEEEbiography}[{\includegraphics[width=1in,height=1.25in,clip,keepaspectratio]{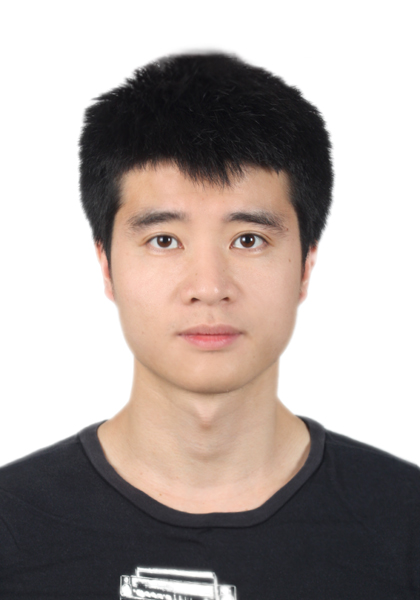}}]{Yang Yang}
is working towards the PhD degree with the National Key Lab for Novel Software Technology, the Department of Computer Science $\&$ Technology in Nanjing University, China. His research interests lie primarily in machine learning and data mining, including heterogeneous learning, model reuse, and incremental mining.
\end{IEEEbiography}
\vspace{-1.2cm}
\begin{IEEEbiography}[{\includegraphics[width=1in,height=1.25in,clip,keepaspectratio]{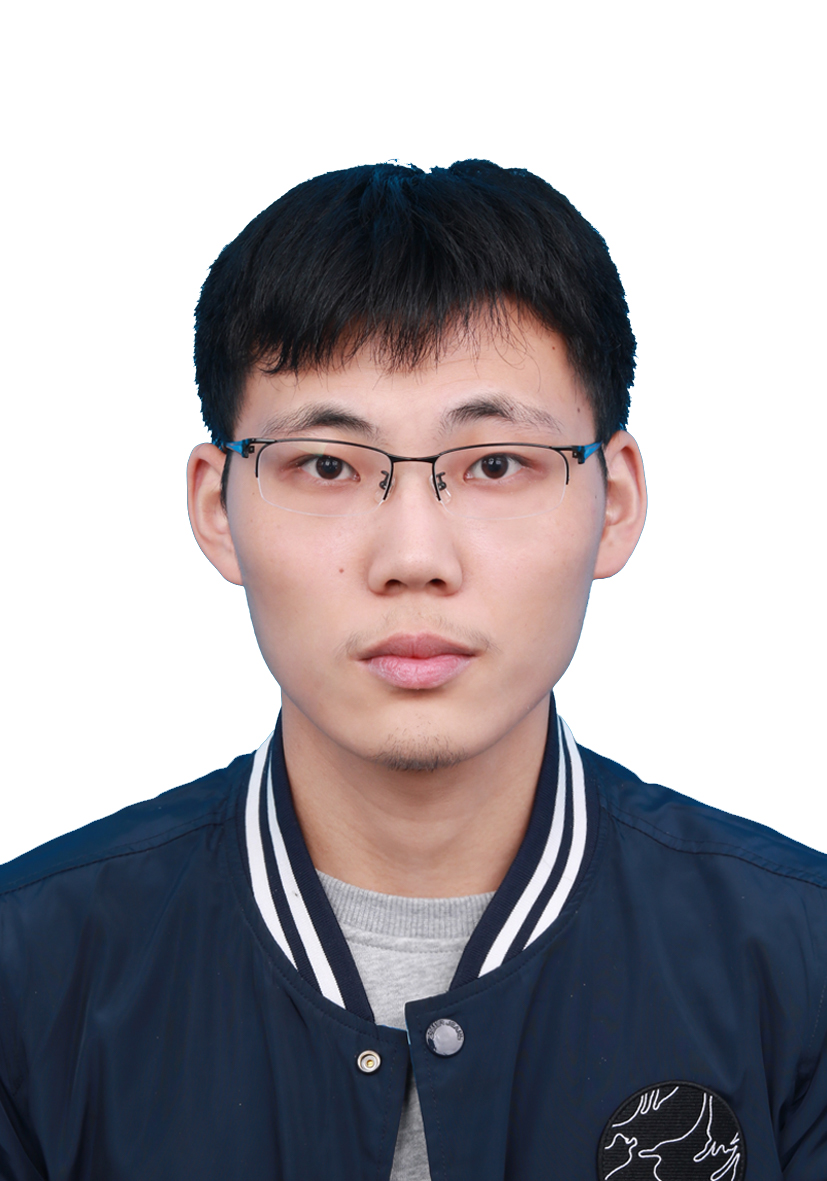}}]{Zhao-Yang Fu}
is working towards the M.Sc. degree with the National Key Lab for Novel Software Technology, the Department of Computer Science $\&$ Technology in Nanjing University, China. His research interests lie primarily in machine learning and data mining, including multi-modal learning.
\end{IEEEbiography}
\vspace{-1.2cm}
\begin{IEEEbiography}[{\includegraphics[width=1in,height=1.25in,clip,keepaspectratio]{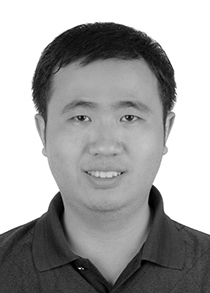}}]{De-Chuan Zhan}
received the Ph.D. degree in computer science, Nanjing University, China in 2010. At the same year, he became a faculty member in the Department of Computer Science and Technology at Nanjing University, China. He is currently an Associate Professor with the Department of Computer Science and Technology at Nanjing University. His research interests are mainly in machine learning, data mining and mobile intelligence. He has published over 20 papers in leading international journal/conferences. He serves as an editorial board member of IDA and IJAPR, and serves as SPC/PC in leading conferences such as IJCAI, AAAI, ICML, NIPS, etc.
\end{IEEEbiography}
\vspace{-1.2cm}
\begin{IEEEbiography}[{\includegraphics[width=1in,height=1.25in,clip,keepaspectratio]{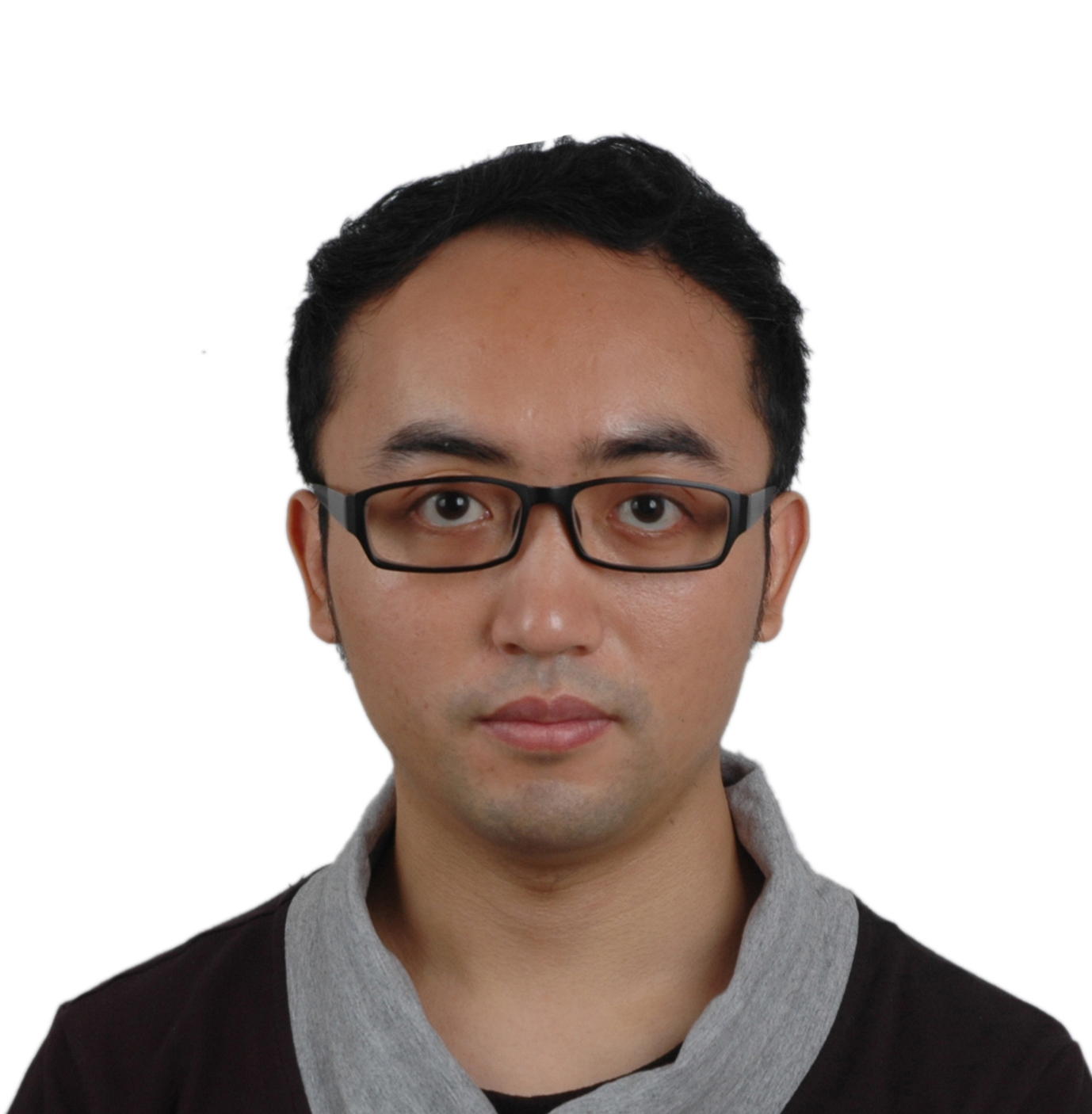}}]{Zhi-Bin Liu}
received the Ph.D. degree and M.S. degree in control science and engineering from Tsinghua Universtiy, Beijing, China, in 2010, and the B.S. degree in automatic control engineering from Central South University, Changsha, China, in 2004. His research interests are in big data minning, machine learning, AI, NLP, computer vision, information fusion and etc.
\end{IEEEbiography}
\vspace{-1.2cm}
\begin{IEEEbiography}[{\includegraphics[width=1in,height=1.25in,clip,keepaspectratio]{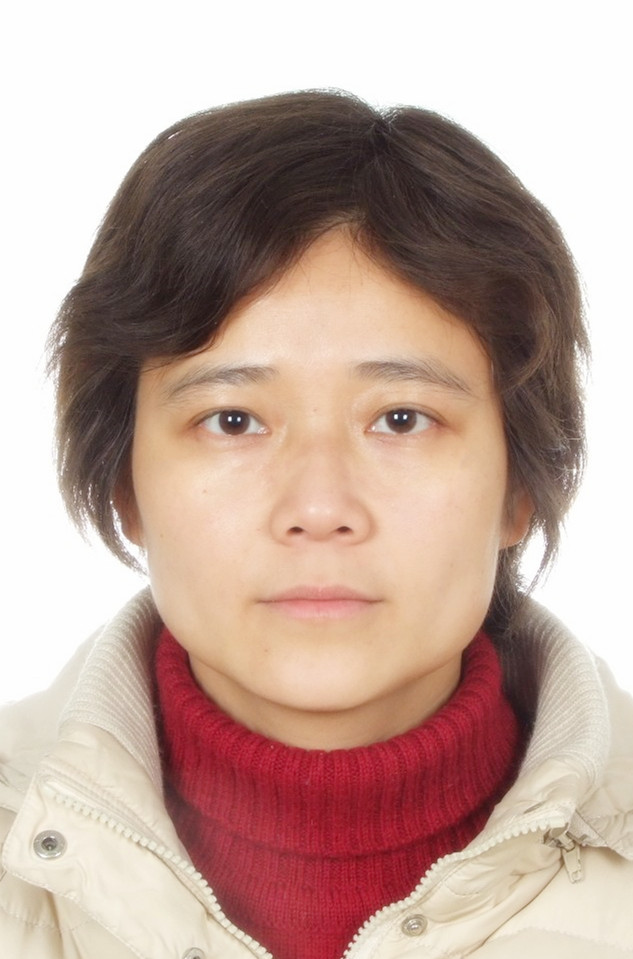}}]{Yuan Jiang}
 received the PhD degree in computer science from Nanjing University, China, in 2004. At the same year, she became a faculty member in the Department of Computer Science $\&$ Technology at Nanjing University, China and currently is a Professor. She was selected in the Program for New Century Excellent talents in University, Ministry of Education in 2009. Her research interests are mainly in artificial intelligence, machine learning, and data mining. She has published over 50 papers in leading international/national journals and conferences.
\end{IEEEbiography}


% % insert where needed to balance the two columns on the last page with
% %\newpage


\end{document}